\documentclass{article} 

\PassOptionsToPackage{sort}{natbib}
\PassOptionsToPackage{svgnames}{xcolor}
\usepackage{iclr2027_conference,times}
\usepackage{type1cm} 

\usepackage[T1]{fontenc}
\usepackage{amsmath}
\usepackage{booktabs}
\usepackage{array} 
\usepackage{algorithm}
\usepackage{algpseudocode}
\usepackage{graphicx}
\usepackage[colorlinks=true, linkcolor=Navy, citecolor=Navy]{hyperref} 
\usepackage{microtype}
\usepackage{url}
\usepackage{csquotes}
 \usepackage{threeparttable}
\usepackage[capitalise]{cleveref} 
\usepackage{tikz}
\usetikzlibrary{arrows.meta, positioning, patterns, calc, fit, backgrounds}

\definecolor{figblue}{HTML}{0072B2}
\definecolor{figgreen}{HTML}{009E73}
\definecolor{figsky}{HTML}{56B4E9}
\definecolor{figpink}{HTML}{CC79A7}
\definecolor{figamber}{HTML}{E69F00}
\definecolor{figvermillion}{HTML}{D55E00}
\definecolor{figgrey}{HTML}{666666}
\definecolor{figviolet}{HTML}{6A51A3}
\definecolor{figgraydark}{HTML}{4A4A4A}
\definecolor{figgraylight}{HTML}{9A9A9A}
\definecolor{figframe}{HTML}{A9B5C3}
\definecolor{figmuted}{HTML}{68737F}

\definecolor{figexit1}{HTML}{84BCDB}
\definecolor{figexit2}{HTML}{5FA8D3}
\definecolor{figexit3}{HTML}{3E8EC4}
\definecolor{figexit4}{HTML}{2270B1}
\definecolor{figexit5}{HTML}{0D549E}
\definecolor{figexit6}{HTML}{08306B}
\definecolor{figcontext1}{HTML}{6BAED6}
\definecolor{figcontext2}{HTML}{4292C6}
\definecolor{figcontext3}{HTML}{2171B5}
\definecolor{figcontext4}{HTML}{08519C}

\colorlet{figcdb}{figgreen}
\colorlet{figcdbnorefill}{figblue}
\colorlet{figbaseline}{figgraydark}
\colorlet{figbound}{figgraylight}

\colorlet{figbluewash}{figblue!20}
\colorlet{figbluefill1}{figblue!30}
\colorlet{figbluefill2}{figblue!48}
\colorlet{figbluefill3}{figblue!66}
\colorlet{figgreenwash}{figgreen!20}
\colorlet{figgreenfill1}{figgreen!30}
\colorlet{figgreenfill2}{figgreen!48}
\colorlet{figgreenfill3}{figgreen!66}
\colorlet{figvioletwash}{figviolet!20}
\colorlet{figvioletfill1}{figviolet!30}
\colorlet{figvioletfill2}{figviolet!48}
\colorlet{figvioletfill3}{figviolet!66}
\colorlet{figamberwash}{figamber!20}
\colorlet{figamberfill1}{figamber!30}
\colorlet{figamberfill2}{figamber!48}
\colorlet{figamberfill3}{figamber!66}
\colorlet{figvermillionwash}{figvermillion!20}
\colorlet{figvermillionfill1}{figvermillion!30}
\colorlet{figvermillionfill2}{figvermillion!48}
\colorlet{figvermillionfill3}{figvermillion!66}
\colorlet{figframewash}{figframe!20}
\colorlet{figframefill1}{figframe!30}
\colorlet{figframefill2}{figframe!48}
\colorlet{figframefill3}{figframe!66}

\colorlet{figfillink}{black!85}

\title{Depth-adaptive Inference of Looped Language Models via Continuous Depth Batching}

\author{%
  Kristian Schwethelm$^{1}$ \quad
  Daniel R\"uckert$^{1,2,3}$ \quad
  Georgios Kaissis$^{4}$\\
  \normalfont\small $^{1}$Chair for AI in Healthcare and Medicine, Technical University of Munich (TUM), Germany\\
  \normalfont\small $^{2}$Department of Computing, Imperial College London, UK\\
  \normalfont\small $^{3}$Munich Center for Machine Learning (MCML), Germany\\
  \normalfont\small $^{4}$Hasso Plattner Institute for Digital Engineering, University of Potsdam, Germany%
}

\iclrfinalcopy

\begin{document}

\maketitle
\lhead{Preprint.}

\begin{abstract}
A main promise of looped language models is \emph{depth-adaptive inference}.
By looping a block of shared layers a variable number of times, the model can use less compute for \enquote{easy} tokens and more for \enquote{hard} ones.
However, tokens with different numbers of loops cannot share a uniform forward pass and therefore cannot be handled by standard batching systems such as vLLM.
The practical value of depth-adaptive inference thus hinges on whether batching can be made efficient.
We introduce the first efficient method for depth-adaptive looped LMs via \emph{continuous depth batching (CDB)}, which forms new batches \emph{between} loop steps.
Our method dynamically schedules looped and non-looped parts of the architecture, manages looped KV-caching, and predicts which tokens will exit the loop in advance so it can prepare batches asynchronously.
Experiments on Ouro 1.4B and Huginn 3.5B show that fully looped architectures are best suited to depth-adaptive inference, as large non-looped layers outside the recurrent core (e.g., token embedding, LM head, and unshared transformer blocks) slow down and complicate scheduling.
Overall, CDB realizes up to 99\% of the estimated maximum speedup available, leaving further gains primarily dependent on model architecture and exit behavior.
\end{abstract}

\section{Introduction}

Batching, i.e., processing multiple sequences in one forward pass, is central to large language model (LLM) inference.
Online APIs, reinforcement learning, and benchmarking require processing many sequences at once.
Batching is especially important for autoregressive decoding, which generates responses one token at a time.
Without batching, each decode step loads the model weights from memory to process a single token.
Weight loading often takes much longer than the computation, leaving compute units underutilized; the forward pass is \emph{memory-bandwidth bound}~\citep{anthony2024codesign}.
By processing multiple sequences, batching reuses the loaded weights and turns small matrix multiplications into larger, more efficient ones.
With a large enough batch, the forward pass can even become \emph{compute bound}, saturating the compute units while weight-loading time becomes negligible.

Hardware constraints and inference systems also increasingly influence LLM architecture design~\citep{bian2026scaling,anthony2024codesign}.
For example, layer dimensions are chosen based on accelerator utilization, modern attention layers reduce KV-cache memory footprint, and native multi-token prediction enables self-speculative decoding~\citep{deepseekai2025deepseekv3technicalreport,pmlr-v235-gloeckle24a}.

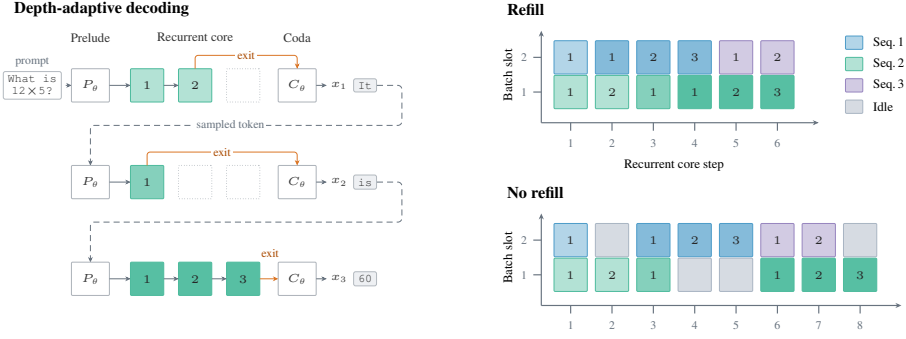
\begin{figure}[t]
  \centering
  \colorlet{slotrule}{figframe}
  \colorlet{mutedlabel}{figmuted}
  \colorlet{sone}{figblue}
  \colorlet{stwo}{figgreen}
  \colorlet{sthree}{figviolet}
  \colorlet{exitc}{figvermillion}
  \resizebox{.86\linewidth}{!}{%
  \begin{tikzpicture}[
    x=1.15932cm, y=1.15932cm,
    cell/.style={draw=slotrule, line width=0.289pt, rounded corners=0.5785pt, minimum width=0.5785cm, minimum height=0.5785cm, inner sep=0pt},
    s1a/.style={cell, draw=sone!70!white, fill=figbluefill1},
    s1b/.style={cell, draw=sone!70!white, fill=figbluefill2},
    s2a/.style={cell, draw=stwo!70!white, fill=figgreenfill1},
    s2b/.style={cell, draw=stwo!70!white, fill=figgreenfill2},
    s2c/.style={cell, draw=stwo!70!white, fill=figgreenfill3},
    s3a/.style={cell, draw=sthree!70!white, fill=figvioletfill1},
    waste/.style={cell, fill=figframefill2},
    skip/.style={cell, draw=mutedlabel!60!white, densely dotted, fill=none},
    mbox/.style={cell, draw=mutedlabel!55!white, minimum width=0.66cm, fill=white},
    ptitle/.style={font=\fontsize{8.099pt}{8.099pt}\selectfont\bfseries, anchor=west},
    legend/.style={font=\fontsize{6.364pt}{6.364pt}\selectfont},
    lab/.style={font=\fontsize{6.364pt}{6.364pt}\selectfont},
    hdr/.style={font=\fontsize{6.364pt}{6.364pt}\selectfont, text=black!80},
    axnum/.style={font=\fontsize{5.785pt}{5.785pt}\selectfont, text=mutedlabel},
    celltext/.style={font=\fontsize{5.785pt}{5.785pt}\selectfont, text=figfillink},
    boxtext/.style={font=\fontsize{5.785pt}{5.785pt}\selectfont, text=black!80},
    rule/.style={draw=mutedlabel, line width=0.5785pt},
    axarrow/.style={rule, -{Stealth[length=3.008pt, width=2.545pt]}},
    flow/.style={draw=mutedlabel, line width=0.35pt, -{Stealth[length=2.4pt, width=2.1pt]}},
    exitflow/.style={draw=exitc!85!white, line width=0.45pt, -{Stealth[length=2.6pt, width=2.3pt]}},
    exitlab/.style={font=\fontsize{5.785pt}{5.785pt}\selectfont, text=exitc!80!black, fill=white, inner sep=0.9pt},
    feed/.style={draw=mutedlabel, line width=0.35pt, densely dashed, -{Stealth[length=2.4pt, width=2.1pt]}},
    feedlab/.style={font=\fontsize{5.785pt}{5.785pt}\selectfont, text=mutedlabel, fill=white, inner sep=0.9pt},
    tokchip/.style={draw=mutedlabel!50!white, rounded corners=1pt, fill=figframewash, inner xsep=2.2pt, inner ysep=1.6pt, font=\fontsize{5.785pt}{5.785pt}\selectfont\ttfamily, text=black!85}
  ]
    \newcommand{\cw}{0.62}
    \newcommand{\gut}{0.55}
    \newcommand{\ox}{7.7}
    \newcommand{\oy}{0}
    \newcommand{\slotone}{-0.52}
    \newcommand{\slottwo}{0}
    \newcommand{\axx}{\ox+\gut-0.44}
    \newcommand{\axy}{-0.98}
    \newcommand{\stepcell}[4]{%
      \node[#3] at ({\ox+\gut+(#1-1)*\cw}, {\oy+#2}) {};%
      \node[celltext] at ({\ox+\gut+(#1-1)*\cw}, {\oy+#2}) {#4};%
    }
    \newcommand{\panelaxes}[2]{%
      \node[ptitle] at ({\axx-0.62}, {\oy+0.72}) {#1};%
      \draw[axarrow] ({\axx}, {\oy+\axy}) -- ({\axx}, {\oy+\slottwo+0.40});%
      \foreach \s/\sy in {1/\slotone, 2/\slottwo} {%
        \draw[rule] ({\axx}, {\oy+\sy}) -- ({\axx-0.12}, {\oy+\sy});%
        \node[axnum] at ({\axx-0.26}, {\oy+\sy}) {\s};%
      }%
      \node[lab, rotate=90] at ({\axx-0.50}, {\oy+0.5*(\slotone+\slottwo)}) {Batch slot};%
      \draw[axarrow] ({\axx}, {\oy+\axy}) -- ({\ox+\gut+(#2-1)*\cw+0.62}, {\oy+\axy});%
      \foreach \i in {1,...,#2} {%
        \draw[rule] ({\ox+\gut+(\i-1)*\cw}, {\oy+\axy}) -- ({\ox+\gut+(\i-1)*\cw}, {\oy+\axy-0.12});%
        \node[axnum] at ({\ox+\gut+(\i-1)*\cw}, {\oy+\axy-0.30}) {\i};%
      }%
    }
    \newcommand{\passlabel}[1]{%
      \node[lab] at ({\ox+\gut+0.5*(#1-1)*\cw}, {\oy+\axy-0.64}) {Recurrent core step};%
    }

    \newcommand{\lpx}{1.05}
    \newcommand{\lra}{1.90}
    \newcommand{\lrb}{2.62}
    \newcommand{\lrc}{3.34}
    \newcommand{\lcx}{4.15}
    \newcommand{\lfx}{5.72}
    \newcommand{\rowa}{0.80}
    \newcommand{\rowb}{-0.65}
    \newcommand{\rowc}{-2.10}
    \newcommand{\reccell}[4]{\node[#3] at (#1,#2) {};\node[celltext] at (#1,#2) {#4};}
    \newcommand{\lprowbase}[1]{%
      \node[mbox] at (\lpx,#1) {};\node[boxtext] at (\lpx,#1) {$P_\theta$};%
      \node[mbox] at (\lcx,#1) {};\node[boxtext] at (\lcx,#1) {$C_\theta$};%
      \draw[flow] ({\lpx+0.285},#1) -- ({\lra-0.25},#1);%
    }
    \newcommand{\lpexit}[2]{%
      \draw[exitflow, rounded corners=1.5pt] (#1,{#2+0.25}) -- (#1,{#2+0.45}) -- (\lcx,{#2+0.45}) -- (\lcx,{#2+0.25});%
      \node[exitlab] at ({0.5*(#1+\lcx)},{#2+0.45}) {exit};%
    }
    \newcommand{\lpout}[4]{%
      \draw[flow] ({\lcx+0.285},#1) -- ({\lcx+0.44},#1);%
      \node[boxtext, anchor=west, inner sep=1.5pt] at ({\lcx+0.46},#1) {$x_#2$};%
      \node[tokchip, anchor=west] (tok#2) at ({\lcx+0.84},#1) {#4};%
      \ifx#3\relax\else%
        \draw[feed, rounded corners=2pt] (tok#2.east) -- (\lfx,#1) -- (\lfx,{#3+0.85}) -- (\lpx,{#3+0.85}) -- (\lpx,{#3+0.25});%
      \fi%
    }

    \node[ptitle] at (-0.20, 1.94) {Depth-adaptive decoding};
    \node[hdr] at (\lpx, 1.52) {Prelude};
    \node[hdr] at (\lrb, 1.52) {Recurrent core};
    \node[hdr] at (\lcx, 1.52) {Coda};

    \node[tokchip, fill=white, align=center, anchor=east] (promptchip) at (0.62,\rowa) {What is\\12$\times$5?};
    \node[feedlab, anchor=south] at ($(promptchip.north)+(0,0.03)$) {prompt};
    \draw[flow] (0.68,\rowa) -- ({\lpx-0.285},\rowa);
    \lprowbase{\rowa}
    \reccell{\lra}{\rowa}{s2a}{$1$}\reccell{\lrb}{\rowa}{s2a}{$2$}\reccell{\lrc}{\rowa}{skip}{}
    \draw[flow] ({\lra+0.25},\rowa) -- ({\lrb-0.25},\rowa);
    \lpexit{\lrb}{\rowa}
    \lpout{\rowa}{1}{\rowb}{It}
    \node[feedlab] at (3.15,{\rowb+0.85}) {sampled token};

    \lprowbase{\rowb}
    \reccell{\lra}{\rowb}{s2b}{$1$}\reccell{\lrb}{\rowb}{skip}{}\reccell{\lrc}{\rowb}{skip}{}
    \lpexit{\lra}{\rowb}
    \lpout{\rowb}{2}{\rowc}{is}

    \lprowbase{\rowc}
    \reccell{\lra}{\rowc}{s2c}{$1$}\reccell{\lrb}{\rowc}{s2c}{$2$}\reccell{\lrc}{\rowc}{s2c}{$3$}
    \draw[flow] ({\lra+0.25},\rowc) -- ({\lrb-0.25},\rowc);
    \draw[flow] ({\lrb+0.25},\rowc) -- ({\lrc-0.25},\rowc);
    \draw[exitflow] ({\lrc+0.25},\rowc) -- ({\lcx-0.285},\rowc);
    \node[exitlab] at ({0.5*(\lrc+\lcx)},{\rowc+0.38}) {exit};
    \lpout{\rowc}{3}{\relax}{60}

    \renewcommand{\oy}{1.22}
    \panelaxes{Refill}{6}
    \passlabel{6}
    \stepcell{1}{\slottwo}{s1a}{$1$}\stepcell{2}{\slottwo}{s1b}{$1$}\stepcell{3}{\slottwo}{s1b}{$2$}
    \stepcell{4}{\slottwo}{s1b}{$3$}\stepcell{5}{\slottwo}{s3a}{$1$}\stepcell{6}{\slottwo}{s3a}{$2$}
    \stepcell{1}{\slotone}{s2a}{$1$}\stepcell{2}{\slotone}{s2a}{$2$}\stepcell{3}{\slotone}{s2b}{$1$}
    \stepcell{4}{\slotone}{s2c}{$1$}\stepcell{5}{\slotone}{s2c}{$2$}\stepcell{6}{\slotone}{s2c}{$3$}

    \renewcommand{\oy}{-1.52}
    \panelaxes{No refill}{8}
    \stepcell{1}{\slottwo}{s1a}{$1$}\stepcell{2}{\slottwo}{waste}{}
    \stepcell{1}{\slotone}{s2a}{$1$}\stepcell{2}{\slotone}{s2a}{$2$}
    \stepcell{3}{\slottwo}{s1b}{$1$}\stepcell{4}{\slottwo}{s1b}{$2$}\stepcell{5}{\slottwo}{s1b}{$3$}
    \stepcell{3}{\slotone}{s2b}{$1$}\stepcell{4}{\slotone}{waste}{}\stepcell{5}{\slotone}{waste}{}
    \stepcell{6}{\slottwo}{s3a}{$1$}\stepcell{7}{\slottwo}{s3a}{$2$}\stepcell{8}{\slottwo}{waste}{}
    \stepcell{6}{\slotone}{s2c}{$1$}\stepcell{7}{\slotone}{s2c}{$2$}\stepcell{8}{\slotone}{s2c}{$3$}

    \newcommand{\legx}{12.5}
    \newcommand{\legendrow}[3]{%
      \node[#2, minimum width=0.3008cm, minimum height=0.2314cm] at (\legx,#1) {};%
      \node[legend, anchor=west] at ({\legx+0.18},#1) {#3};%
    }
    \legendrow{1.44}{s1a}{Seq.\,1}
    \legendrow{1.12}{s2a}{Seq.\,2}
    \legendrow{0.80}{s3a}{Seq.\,3}
    \legendrow{0.48}{waste}{Idle}
  \end{tikzpicture}%
  }
  \caption{\emph{Left:} the looped LM decodes three tokens, each running the prelude and coda once and the recurrent core a different number of times.
  \emph{Right:} three sequences under CDB with two concurrent batch slots.
  Refill fills slots freed by an exit with a new token, keeping a larger batch size and finishing in 6 forward passes.
  No refill reduces the batch size on token exits, needing 8 forward passes.}
  \label{fig:cdb-slot-grid}
\end{figure}

Looped language models, however, have received little attention from the inference perspective, despite offering several potential advantages.
A looped LM repeatedly applies a block of shared layers, called the \emph{recurrent core}, to refine the latent representation before predicting a token.
Sharing layers reduces parameter memory relative to an equally deep standard transformer and also makes it possible to reuse KV states across depth~\citep{geiping2025huginn}.
The key advantage we focus on is \emph{depth-adaptive decoding}, which allows tokens to exit the recurrent core after different depths (\cref{fig:cdb-slot-grid}, left)~\citep{graves2016adaptive,dehghani2018universal}.
The model spends fewer loops and less compute on \enquote{easy} tokens, and more loops on \enquote{hard} ones, improving the compute--accuracy trade-off~\citep{zhu2025ouro}.

However, the compute savings from depth-adaptive decoding do not automatically translate into faster inference.
The central challenge is that tokens with different numbers of loops cannot share a uniform forward pass and therefore cannot be handled by standard batching.
A promising solution is \emph{continuous depth batching (CDB)}~\citep{bae2024relaxed}, a depth-wise scheduler that forms new batches between loop steps \emph{within} the forward pass.
CDB removes tokens from the loop as they exit and refills the freed slots with new tokens to keep batch size large (\cref{fig:cdb-slot-grid}, right).
However, this makes scheduling much more complex and a naive implementation can incur more overhead than early exit saves, leaving standard fixed-depth batching faster.
In this work, we ask: \emph{Can depth-adaptive looped LMs be served efficiently? Which architectural properties determine their speedup?}

For this, we introduce the first end-to-end CDB implementation, addressing several open challenges.
Looped LMs contain boundary stages outside the shared core that run only once and must be scheduled alongside changing loop batches: the \emph{prelude} processes each token before it enters the loop, and the \emph{coda} predicts the next token after it exits (\cref{fig:cdb-slot-grid}, left).
New tokens for refilling freed batch slots are also not directly available: the next token of an exit needs to pass through coda and prelude and a new request must first complete prompt prefill and prelude.
Furthermore, exit decisions become available only after a loop step finishes, so the scheduler cannot prepare the next recurrent batch in advance, leaving the GPU idle.
To address these challenges, we interleave prefill, prelude, and coda batches with loop steps, allowing new and exiting sequences to provide replacement tokens for refill, and use a lookahead gate to overlap batch preparation with GPU execution.
We implement the original refill scheduler and a simpler \emph{no-refill} mode that only removes tokens (\cref{fig:cdb-slot-grid}, right).

To analyze CDB, we derive a prefill-corrected speedup estimate and a hardware-aware latency model that predicts when refill helps.
We evaluate CDB in offline throughput and online serving benchmarks on two prominent looped architectures.
Our results show that refill is important, especially in the memory-bound regime and for fully looped models such as Ouro~\citep{zhu2025ouro}, whose boundary stages are minimal.
On Ouro, CDB realizes up to 99\% of the estimated maximum speedup available from early exit.
For Huginn~\citep{geiping2025huginn}, costly transformer boundary stages increase the cost of refill and no-refill performs better.
Here, CDB realizes 83 to 96\% of the available speedup.

\textbf{Contributions.}
\begin{enumerate}
\item We enable efficient depth-adaptive inference for looped LMs with the first end-to-end implementation of CDB.
\item We compare theoretical compute savings with wall-clock speed-up by deriving a FLOP-based upper bound and a hardware-aware model of recurrent-step latency.
\item We show how boundary-stage cost, KV cache design, and loop count determine inference efficiency, providing concrete guidance for inference-aware looped-LM architecture design.
\end{enumerate}

\section{Problem Setup}
\label{sec:problem-setup}

The looped-LM architecture is split into a \emph{prelude} $P_\theta$, a shared \emph{recurrent core} $R_\theta$, and a \emph{coda} $C_\theta$.
For token $t$, the recurrent core runs $r_t\leq r_{\max}$ times, while the prelude and coda each run once.
We denote the number of transformer layers in these stages as prelude-core-coda.
For example, a 2-4-2 model executes $2{+}4r_t{+}2$ layers, giving an effective depth of 20 at $r_t{=}4$ loops.
A fully looped 0-4-0 model has no transformer layers outside the core; its boundary stages (i.e., prelude and coda) only contain the embedding layer and LM head.
Within this architecture, depth-adaptive decoding lets $r_t$ vary per token: after each loop, an exit criterion determines whether the token stops looping.
Crucially, each token leaves the core at an \emph{input-dependent} depth, decided \emph{at the end of a loop step}.

This per-token variability creates a problem for batched inference.
Standard \emph{continuous batching (CB)} schedules only at token boundaries: at each tick, active sequences generate one token, completed sequences are removed, and new sequences are admitted~\citep{yu2022orca}.
Because CB treats the complete forward pass as one scheduling unit, every token in a batch must execute the same operations.
With adaptive depth, however, an exited token must run the coda while the remaining tokens may require another loop step.
To support this, the CDB scheduler must operate \emph{within} the forward pass by tracking each token's loop progress individually and routing tokens between stages as they exit.

Overall, CDB must address three main challenges:
(1) \emph{stage heterogeneity}---a CB step is one uniform forward pass, while CDB must schedule prefill, prelude, recurrent core, and coda separately, each at different batch sizes and frequencies;
(2) \emph{missing KV states}---when a token exits at $r_t$, it writes no KV states for deeper loop steps, leaving later tokens without a state to attend to if they loop deeper;
(3) \emph{critical-path exits}---the exit decision is available only \emph{after} step $r$ but determines which tokens enter step $r+1$, so the GPU idles after each loop step until the scheduler has processed the exit.

\section{Related Work}
\label{sec:related}

\textbf{Depth-adaptive looped transformers.}
We focus on methods that decide for each token \emph{individually} when it leaves the loop. 
Here, learned variants attach a small exit head to the recurrent core~\citep{banino2021pondernet,tan2023sparse,zhu2025ouro,fu2025thinkathard,zeng2026ponderlm3,song2026adaponderlm,frey2026adaptive}, while training-free variants compare latent states or output logits between consecutive loop steps and exit once they converge~\citep{geiping2025huginn,kohli2026loop,lee2026sparse}.
In contrast, in Mixture-of-Depths-style methods, tokens exit in fixed-sized \emph{groups}~\citep{raposo2024mixture,bae2025mixture,mohtashami2023cotformer}, while in sequence-level methods, \emph{all} tokens exit at the same depth~\citep{jeddi2026loopformer,prairie2026parcae}.
We study per-token adaptive depth from a \emph{systems} perspective, quantifying how its FLOP savings translate into throughput and latency gains.
Importantly, CDB does not change the model's predictions and exit decisions.

\textbf{Depth-adaptive standard transformers.}
Early exit has also been studied for standard transformers, where intermediate layers can predict the next token and stop computation early~\citep{elbayad2020depth,schuster2022calm,elhoushi2024layerskip,yoo2026adeptadaptivedynamicearlyexit}.
However, efficient batched inference remains difficult because their unshared layers prevent refill across depths and leave missing KV states that must be recomputed or approximated when needed~\citep{delcorro2023skipdecode,liu2025drex,bae2023free,chen2023eellm,shen2026river}.
We study CDB only for looped LMs, as their shared recurrent core naturally allows cross-depth refill and KV sharing without recomputation.

\textbf{Continuous depth batching.}
\citet{bae2024relaxed} introduced CDB for looped LMs theoretically and estimated its throughput speedup through simulation. \citet{bae2025mixture} later evaluated CDB in ablation experiments, excluding prefill, scheduling, and KV-cache updates.
In contrast, we include all of these costs and provide the first end-to-end implementation and evaluation of CDB.

\section{Method: Continuous Depth Batching}
\label{sec:method}

\begin{figure}[t]
  \centering
  \resizebox{\linewidth}{!}{%
  \begin{tikzpicture}[
    font=\scriptsize,
    box/.style={draw=figframe, rounded corners=2pt, line width=0.4pt, align=center, inner sep=3pt},
    sched/.style={box, fill=figbluefill1, draw=figblue!68!white, text width=2.5cm, minimum height=0.95cm},
    schedbox/.style={rounded corners=3pt, draw=figblue!45!white, densely dashed, line width=0.5pt},
    schedtitle/.style={font=\scriptsize\bfseries, text=figblue!60!black},
    stage/.style={box, fill=figgreenfill1, draw=figgreen!72!white, text width=2.0cm, minimum height=0.95cm},
    core/.style={stage, fill=figgreenfill2, text width=2.1cm},
    extern/.style={box, fill=white, draw=figblue!40!white, text width=1.35cm, minimum height=0.95cm},
    arrow/.style={-{Stealth[length=1.9mm]}, line width=0.5pt, draw=figmuted},
    loop/.style={-{Stealth[length=1.9mm]}, line width=0.6pt, draw=figmuted},
    feed/.style={-{Stealth[length=1.7mm]}, line width=0.45pt, draw=figblue!60!black, densely dashed},
    slab/.style={font=\tiny, text=figmuted, fill=white, inner sep=1.2pt},
    flab/.style={font=\tiny, text=figblue!58!black, fill=white, inner sep=1.2pt},
    ptitle/.style={font=\fontsize{8.5}{9.2}\selectfont\bfseries, text=black, anchor=west},
    kcell/.style={draw, rounded corners=1pt, line width=0.4pt, minimum width=0.5cm, minimum height=0.5cm, inner sep=0pt},
    compa/.style={kcell, draw=figgreen!72!white, fill=figgreenfill1},
    compb/.style={kcell, draw=figgreen!72!white, fill=figgreenfill2},
    compc/.style={kcell, draw=figgreen!72!white, fill=figgreenfill3},
    kcopy/.style={kcell, draw=figgreen!60!white, densely dashed, fill=figgreenwash},
    ktitle/.style={font=\scriptsize\bfseries},
    klab/.style={font=\scriptsize, text=figmuted},
    kstate/.style={font=\scriptsize, text=black!60},
  ]
    \node[ptitle] at (-0.85,2.62) {Stage-wise scheduling};
    \tikzset{
      tracebox/.style={box, text width=1.2cm, minimum height=0.78cm, font=\scriptsize},
      tracecore/.style={tracebox, fill=figgreenfill2, draw=figgreen!72!white},
      tracecoda/.style={tracebox, fill=figbluefill1, draw=figblue!68!white},
      tracepre/.style={tracecoda},
      rowlabel/.style={anchor=east, align=right, font=\scriptsize},
      note/.style={font=\tiny, text=figmuted, align=center},
    }
    \node[rowlabel,anchor=west] at (1.5,1.8) {\textbf{Refill} ($K{=}1$)};
    \node[rowlabel,anchor=west] at (1.5,0.54) {\textbf{Refill} ($K{=}16$)};
    \node[rowlabel,anchor=west] at (1.5,-0.71) {\textbf{No-refill}};
    \node[tracecore] (start) at (0,-0.05) {Core\\$B{=}64$};
    \coordinate (split) at (1.5,-0.05);
    \draw[line width=0.5pt,draw=figmuted] (start.east) --
      node[note,above]{1 exit} (split);
    \fill[figmuted] (split) circle (1.2pt);

    \node[tracecore] (n1) at (2.85,-1.32) {Core\\$B{=}63$};
    \node[tracecore] (n2) at (5.2,-1.32) {Core\\$B{=}34$};
    \node[tracecoda] (n3) at (7.55,-1.32) {Coda\\$B{=}64$};
    \node[tracepre] (n4) at (9.9,-1.32) {Prelude\\$B{=}64$};
    \draw[arrow] (n3) -- (n4);
    \draw[arrow] (n1) -- node[note,above,yshift=2pt]{29 exit} (n2);
    \draw[arrow] (n2) -- node[note,above,yshift=2pt]{34 exit} (n3);

    \node[tracecoda] (a1) at (2.85,1.22) {Coda\\$B{=}1$};
    \node[tracepre] (a2) at (5.2,1.22) {Prelude\\$B{=}1$};
    \node[tracecore] (a3) at (7.55,1.22) {Core\\$B{=}64$};
    \draw[arrow] (a1) -- (a2);
    \draw[arrow] (a2) -- (a3);

    \node[tracecore] (b1) at (2.85,-0.05) {Core\\$B{=}63$};
    \node[tracecoda] (b2) at (5.2,-0.05) {Coda\\$B{=}30$};
    \node[tracepre] (b3) at (7.55,-0.05) {Prelude\\$B{=}30$};
    \node[tracecore] (b4) at (9.9,-0.05) {Core\\$B{=}64$};
    \draw[arrow] (b1) -- node[note,above,yshift=2pt]{29 exit} (b2);
    \draw[arrow] (b2) -- (b3);
    \draw[arrow] (b3) -- (b4);
    \draw[arrow] (split) |- (n1.west);
    \draw[arrow] (split) |- (a1.west);
    \draw[arrow] (split) -- (b1.west);

    \node[ptitle] at (11.8,2.62) {Depth-aware KV cache};
    \begin{scope}[shift={(12.85,-0.33)}]
      \newcommand{\kv}[4]{\node[#3] at (#1,#2) {};\node[kstate] at (#1,#2) {#4};}

      \node[ktitle] at (0.58,1.9) {Last-exited};
      \kv{0}{0}{compa}{$1$}\kv{0.58}{0}{compb}{$1$}\kv{1.16}{0}{compc}{$1$}
      \kv{0}{0.58}{compa}{$2$}\kv{0.58}{0.58}{kcopy}{$1$}\kv{1.16}{0.58}{compc}{$2$}
      \kv{0}{1.16}{kcopy}{$2$}\kv{0.58}{1.16}{kcopy}{$1$}\kv{1.16}{1.16}{compc}{$3$}
      \foreach \s/\sy in {1/0, 2/0.58, 3/1.16} {\node[klab] at (-0.44,\sy) {\s};}

      \node[ktitle] at (3.33,0.7) {Shared};
      \kv{2.75}{0}{compa}{$2$}\kv{3.33}{0}{compb}{$1$}\kv{3.91}{0}{compc}{$3$}
      \node[klab] at (2.31,0) {1};
      \node[klab, rotate=90] at (-0.82,0.58) {Depth slot};

      \foreach \t/\tx in {1/0, 2/0.58, 3/1.16} {
        \node[klab] at (\tx,-0.42) {$x_\t$};
        \node[klab] at ({\tx+2.75},-0.42) {$x_\t$};
      }

      \node[compb, minimum width=0.3cm, minimum height=0.22cm] at (0.7,-1.05) {};
      \node[klab, anchor=west] at (0.86,-1.05) {computed};
      \node[kcopy, minimum width=0.3cm, minimum height=0.22cm] at (2.35,-1.05) {};
      \node[klab, anchor=west] at (2.51,-1.05) {copy-on-exit};
    \end{scope}
  \end{tikzpicture}%
  }
  \caption{%
    \emph{Left:} three schedules after a single token exits the loop (core); $B$ denotes the stages' batch size.
Refill interleaves coda and prelude with recurrent-core steps, while the minimum coda batch size $K$ determines how many exits accumulate before coda runs.
No-refill lets the recurrent batch shrink and runs coda after all tokens have exited.
    \emph{Right:} depth-aware KV cache designs for three tokens that exit at depths $(r_1, r_2,r_3)=(2,1,3)$.
    Each cell shows the loop step whose state is stored.
  }
  \label{fig:cdb-overview}
\end{figure}
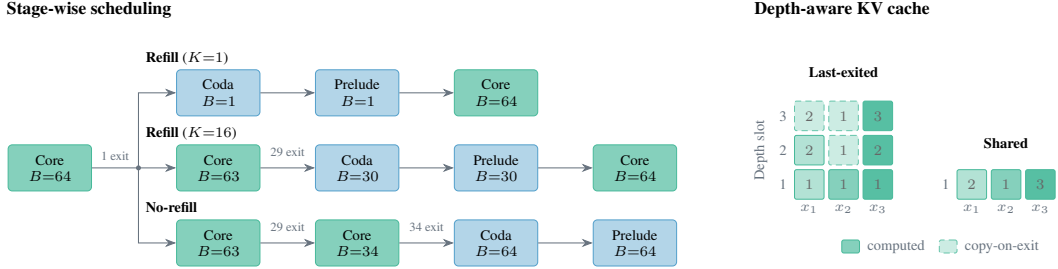

\subsection{Stage-wise Scheduling}
\label{sec:depth-scheduled-decode}

We conceptually split each request's path from prompt to completed response into four queues: inactive requests waiting for prefill, decode tokens awaiting the prelude, recurrent items ready for their next loop step, and exited tokens awaiting the coda.
These queues correspond to the four stages of looped inference, which can be batched and run separately.
In principle, the scheduler can execute them in any order, allowing prefill, prelude, recurrent, and coda batches to be interleaved.

Based on this queuing system, we implement two execution modes that differ in how the stages are ordered.
In \emph{no-refill} mode (\cref{alg:no-refill}), stages follow a fixed order: prelude, then recurrent steps until \emph{all} tokens have exited, then coda.
Exited tokens are removed from the recurrent batch, reducing batch size and thus compute on subsequent loop steps.
In \emph{refill} mode (\cref{alg:refill}), the scheduler orders the stages dynamically so that freed slots can be filled by new tokens, keeping batch size high.

For refill, we use a priority rule to determine the dynamic order.
We let prelude run directly after prefill or coda, as there is no routing decision between these stages.
Among the remaining stages, coda has highest priority because it predicts next tokens for refill and finishing requests.
Prefill comes next, since admitting prompts early allows for larger decode batches.
Otherwise, we run the recurrent core.
Without an additional constraint, this priority system would run the coda after every individual exit.
For models with costly coda layers, repeatedly running them on only a few tokens can outweigh the benefit of refill.
We therefore introduce a minimum coda batch size $K$: coda takes priority only after at least $K$ exits have accumulated.
See implementation details in \cref{app:impl}.

\cref{fig:cdb-overview} (left) illustrates what happens after one token exits a recurrent core batch of 64.
With refill ($K{=}1$), coda and prelude run immediately for the single exited token, so the next recurrent batch refills to $B{=}64$.
With refill ($K{=}16$), the recurrent batch first continues with 63 tokens, allowing more exits to accumulate before coda and prelude run on 30 tokens together.
No-refill waits until the entire recurrent batch has exited and processes all 64 tokens through the coda together.
A larger $K$ therefore makes coda and prelude batches more efficient, but delays refill and temporarily reduces the recurrent batch size, making it less efficient.
Complete scheduling traces are provided in \cref{app:schedule-trace}.

\subsection{Depth-aware KV Cache}
\label{sec:kv-policy}

A depth-indexed KV cache gives each shared layer a separate slot for every loop step, resulting in $r_{\max}$ slots per token.
When a token exits early ($r_t {<} r_{\max}$), its remaining slots stay empty, but later tokens reaching these loop steps need a state to attend to.
We consider two ways to provide these missing states (\cref{fig:cdb-overview}, right), both compatible with modern attention kernels like paged attention~\citep{kwon2023pagedattention} and FlashAttention~\citep{dao2022flashattention}.
A \emph{last-exited cache} copies a token's final KV state into skipped slots upon exit.
This generally preserves accuracy and avoids computing the missing states~\citep{zhu2025ouro}, but retains $r_{\max}$ slots and adds copying overhead.
A \emph{shared cache} instead gives each recurrent layer a single KV slot per token, reducing recurrent-core KV memory by a factor of $r_{\max}$.
Each loop step overwrites this slot, so later tokens always attend to the most recent state of each earlier token, regardless of its exit depth.
This avoids empty slots and copying, but may require specialized training to preserve accuracy~\citep{zhu2025ouro,geiping2025huginn,vendrell2026melt}.

\subsection{Asynchronous Scheduling}
\label{sec:static-async}

\begin{figure}[t]
  \centering
  \input{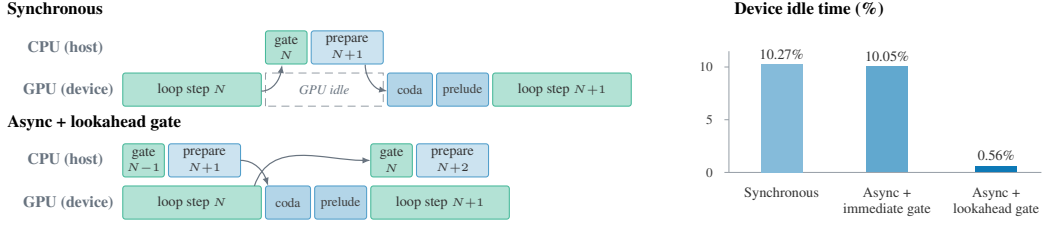}
  \colorlet{mutedlabel}{figmuted}
  \resizebox{\textwidth}{!}{%
  \newcommand{\stepidx}[1]{{\fontsize{4.3pt}{4.7pt}\selectfont $#1$}}
  \begin{tikzpicture}[
    x=1cm, y=1cm,
    lane/.style={font=\fontsize{6pt}{6pt}\selectfont\bfseries, text=mutedlabel, anchor=east},
    rec/.style={draw=figgreen!72!white, fill=figgreenfill1, line width=0.4pt, rounded corners=1pt},
    gate/.style={draw=figgreen!72!white, fill=figgreenfill1, line width=0.4pt, rounded corners=1pt},
    mdl/.style={draw=figblue!68!white, fill=figbluefill1, line width=0.4pt, rounded corners=1pt},
    cpu/.style={draw=figblue!68!white, fill=figbluewash, line width=0.4pt, rounded corners=1pt},
    blk/.style={font=\fontsize{5.4pt}{6pt}\selectfont, text=black!80, align=center},
    tag/.style={font=\fontsize{5.0pt}{5.4pt}\selectfont, text=black!75, align=center},
    ptitle/.style={font=\fontsize{6.5pt}{7pt}\selectfont\bfseries, text=black, anchor=west},
    note/.style={font=\fontsize{5pt}{5.6pt}\selectfont\itshape, text=mutedlabel, align=center},
    axis/.style={draw=mutedlabel!70!white, line width=0.3pt},
    dep/.style={-{Latex[length=1.1mm,width=0.85mm]}, draw=mutedlabel, line width=0.45pt},
  ]
    \begin{scope}[xscale=0.92]
    \node[ptitle] at (-1.8,2.68) {Synchronous};
    \node[lane] at (-0.15,2.18) {CPU (host)};
    \node[lane] at (0,1.62) {GPU (device)};
    \fill[rec] (0,1.4) rectangle (2,1.84);
    \node[blk] at (1.000,1.62) {loop step \stepidx{N}};
    \fill[mdl] (3.82,1.4) rectangle (4.47,1.84);
    \node[tag] at (4.145,1.62) {coda};
    \fill[mdl] (4.53,1.4) rectangle (5.28,1.84);
    \node[tag] at (4.905,1.62) {prelude};
    \fill[rec] (5.34,1.4) rectangle (7.34,1.84);
    \node[blk] at (6.340,1.62) {loop step \stepidx{N{+}1}};
    \fill[gate] (2.06,1.96) rectangle (2.66,2.40);
    \node[blk] at (2.360,2.18) {gate\\\stepidx{N}};
    \fill[cpu] (2.72,1.96) rectangle (3.76,2.40);
    \node[blk] at (3.240,2.18) {prepare\\\stepidx{N{+}1}};
    \draw[axis, densely dashed] (2.06,1.4) rectangle (3.76,1.84);
    \node[note] at (2.910,1.62) {GPU idle};
    \draw[dep] (2.0,1.6) to[out=0,in=-100] (2.3,1.95);
    \draw[dep] (3.5,1.95) to[out=-90,in=170] (3.82,1.6);

    \node[ptitle] at (-1.8,1.18) {Async + lookahead gate};
    \node[lane] at (-0.15,0.68) {CPU (host)};
    \node[lane] at (0,0.12) {GPU (device)};
    \fill[rec] (0,-0.10) rectangle (2,0.34);
    \node[blk] at (1.000,0.12) {loop step \stepidx{N}};
    \fill[mdl] (2.06,-0.10) rectangle (2.71,0.34);
    \node[tag] at (2.385,0.12) {coda};
    \fill[mdl] (2.77,-0.10) rectangle (3.52,0.34);
    \node[tag] at (3.145,0.12) {prelude};
    \fill[rec] (3.58,-0.10) rectangle (5.58,0.34);
    \node[blk] at (4.580,0.12) {loop step \stepidx{N{+}1}};
    \fill[gate] (0,0.46) rectangle (0.6,0.90);
    \node[blk] at (0.300,0.68) {gate\\\stepidx{N{-}1}};
    \fill[cpu] (0.66,0.46) rectangle (1.70,0.90);
    \node[blk] at (1.180,0.68) {prepare\\\stepidx{N{+}1}};
    \fill[gate] (3.58,0.46) rectangle (4.18,0.90);
    \node[blk] at (3.880,0.68) {gate\\\stepidx{N}};
    \fill[cpu] (4.24,0.46) rectangle (5.28,0.90);
    \node[blk] at (4.760,0.68) {prepare\\\stepidx{N{+}2}};
    \draw[dep] (1.9,0.34) to[out=70,in=180] (3.58,0.68);
    \draw[dep] (1.70,0.68) to[out=0,in=110] (2.1,0.34);

    \end{scope}
    \node[ptitle] at (8.0,2.68) {Device idle time (\%)};
    \def\idlebase{0.52}
    \def\idlescale{0.14}
    \def\idlewidth{0.55}
    \draw[axis] (8.05,\idlebase) -- (8.05,{\idlebase+11.5*\idlescale});
    \draw[axis] (8.05,\idlebase) -- (12.20,\idlebase);
    \foreach \v in {0,5,10}{
      \draw[axis] (7.99,{\idlebase+\v*\idlescale}) -- (8.05,{\idlebase+\v*\idlescale});
      \node[tag, anchor=east] at (7.96,{\idlebase+\v*\idlescale}) {\v};
    }
    \fill[figbluefill2] ({8.75-\idlewidth/2},\idlebase) rectangle
      ({8.75+\idlewidth/2},{\idlebase+\idleSync*\idlescale});
    \node[blk] at (8.75,{\idlebase+\idleSync*\idlescale+0.13}) {\idleSyncLabel};
    \fill[figblue!62] ({10.15-\idlewidth/2},\idlebase) rectangle
      ({10.15+\idlewidth/2},{\idlebase+\idleAsyncImmediate*\idlescale});
    \node[blk] at (10.15,{\idlebase+\idleAsyncImmediate*\idlescale+0.13}) {\idleAsyncImmediateLabel};
    \fill[figblue!95] ({11.60-\idlewidth/2},\idlebase) rectangle
      ({11.60+\idlewidth/2},{\idlebase+\idleLookahead*\idlescale});
    \node[blk] at (11.60,{\idlebase+\idleLookahead*\idlescale+0.15}) {\idleLookaheadLabel};
    \node[blk, anchor=north] at (8.75,{\idlebase-0.08}) {Synchronous};
    \node[blk, anchor=north] at (10.15,{\idlebase-0.08}) {Async +\\immediate gate};
    \node[blk, anchor=north] at (11.60,{\idlebase-0.08}) {Async +\\lookahead gate};
  \end{tikzpicture}%
  }
  \caption{%
    \emph{Left:} synchronous scheduling leaves the GPU idle during exit processing and batch preparation.
    Asynchronous lookahead overlaps this work with the GPU step.
    \emph{Right:} asynchronous preparation alone lowers idle time minimally, adding the lookahead gate reduces it to nearly zero.
  }
  \label{fig:cdb-async-overhead}
\end{figure}

CDB introduces frequent CPU interactions between short GPU stages, because CDB schedules every loop and boundary stage separately.
After each stage, the scheduler reads back results, selects the next stage, assembles its batch, and transfers the required inputs to the GPU.
Doing this synchronously results in much GPU idle time.
We therefore prepare batches and transfer data asynchronously while the GPU processes the current batch (\cref{fig:cdb-async-overhead}, left).
We also use acceleration techniques like static, preallocated tensors, CUDA graphs, and batches padded to the next power of two (\cref{app:efficient-execution}).

However, there is a problem with preparing the next batch in advance: the gate output at loop step $r$ determines which tokens leave the batch, but is only available once step $r$ finishes on the GPU.
The CPU must read back the result and build the batch before the GPU can start step $r{+}1$.
End-of-sequence (EOS) tokens pose an analogous problem on the token level, which is usually handled by assuming no EOS and catching it one step late, causing one extra generated token.
We propose a similar idea with a \emph{lookahead gate}.
The exit signal at step $r$ determines whether a token participates in step $r{+}2$ instead of $r{+}1$, giving the CPU a full loop step to prepare the next batch.
Unlike the EOS approach, this does not waste compute, since we decide the exit earlier instead of catching it later.
In \cref{app:early-exit-gating}, we show that the exit decision can be made one step earlier without reducing accuracy.
Overall, our CDB implementation reduces GPU idle time to nearly zero (\cref{fig:cdb-async-overhead}, right).

\section{Efficiency Limits of Adaptive Depth}
\label{sec:cost-model}

We next study what drives efficiency gains for adaptive depth from the hardware perspective. We find that early exits reduce recurrent-core FLOPs, but fewer FLOPs do not always reduce wall-clock time, especially in the memory bound regime. This is where the refill mode of CDB is important.

\textbf{FLOP-based speedup bound.}
Let $F_r$ be the FLOPs of one recurrent-core application for a single token, and $F_0$ the loop-independent FLOPs of the prelude and coda (i.e., the boundary stages).
Fixed-depth decoding requires $F_0 + r_{\max} F_r$ and adaptive-depth decoding $F_0 + r_t F_r$.
For $T$ generated tokens with mean exit depth $\bar d = \frac{1}{T}\sum_{t=1}^{T} r_t$, the FLOP-based upper bound on decode speedup is
\begin{equation}
\label{eq:flop-bound}
S_{\mathrm{dec}}
=
\frac{F_0+r_{\max}F_r}{F_0+\bar d F_r}
\leq
\frac{r_{\max}}{\bar d}.
\end{equation}
For fully looped models whose boundary stages contain only the token embedding and LM head, $F_0{\ll} F_r$ and the bound is close to $r_{\max}/\bar d$.
Non-looped transformer layers increase the boundary-stage cost $F_0$, reducing the savings of exits.
A larger $r_{\max}$ also makes each skipped loop save a smaller fraction of compute.
Thus, model architecture has a substantial impact on the \emph{available} speedup.

The speedup from adaptive-depth is often estimated from decode FLOPs alone, although prefill is unaffected and can add substantial latency.
Since prefill latency depends on the workload and batching configuration, we measure its contribution post hoc.
Let $\alpha$ denote the measured fraction of time spent on prefill.
Adjusting the decode-only bound gives the end-to-end speedup estimate
\begin{equation}
\label{eq:e2e-flop-bound}
S_{\mathrm{e2e}} = \left(\alpha + (1-\alpha)/S_{\mathrm{dec}}\right)^{-1}.
\end{equation}

\begin{figure}[tb]
  \centering
  \hfill
  \includegraphics[width=0.39\linewidth]{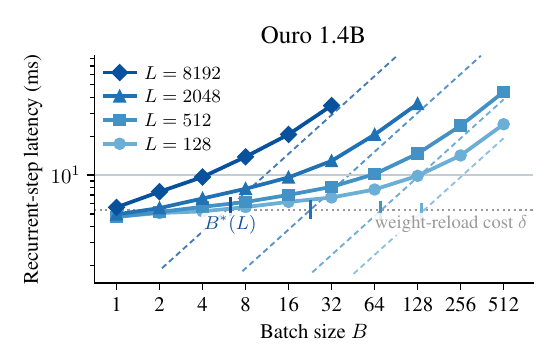}\hfill
  \includegraphics[width=0.39\linewidth]{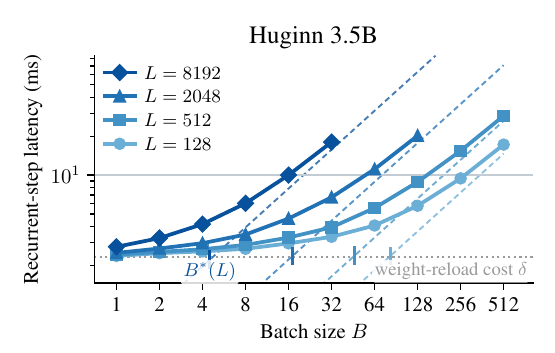}
  \hspace*{\fill}
  \caption{%
    Recurrent step latency versus batch size at context lengths $L$ for Ouro (left) and Huginn (right).
    Dotted lines show weight-reload cost $\delta$, blue ticks mark the saturation batch size $B^*(L)$, and blue dashed lines show linear scaling in the compute bound regime.
    Below $B^*(L)$, shrinking a batch saves little time; above it, smaller sequential batches take roughly as long as one large batch.
  }
  \label{fig:decode-step-latency}
\end{figure}

\textbf{Recurrent batch size and latency.}
CDB refill aims to improve hardware utilization by keeping recurrent batches large, but its benefit depends on how latency scales with batch size.
A recurrent step loads core weights once, while KV cache reads and matrix multiplications scale with batch size, so larger batches reduce the weight loading share.
With $N_b$ batch tokens and $N_c$ context tokens (KV entries), we model the total recurrent step latency as $t(N_c, N_b) = \delta + \gamma\, N_b + \alpha\, N_c$,
where $\delta$ is the fixed weight-reload cost, $\gamma N_b$ the core computation cost, and $\alpha N_c$ the KV cache streaming cost.

\cref{fig:decode-step-latency} shows the recurrent step latency for the two models evaluated below, Ouro 1.4B and Huginn 3.5B.
At small batches, weight loading dominates and latency remains nearly constant, making the step \emph{memory bound}~\citep{williams2009roofline}.
At larger batches, batch-dependent computation and KV-cache streaming dominate, so latency grows approximately linearly with batch size.
The \emph{saturation batch size} $B^\ast$ marks the transition between these regimes.
$B^\ast$ decreases with context length because longer contexts increase the KV-cache streaming cost. See \cref{app:decode-latency} for details.

Thus, \emph{refill works best in the memory-bound regime}.
Here, early exits save FLOPs, but shrinking the batch provides little reduction in latency because weight loading still dominates.
Refill fills the freed slots with new tokens at little additional latency.
In the compute-bound regime, latency decreases as the batch shrinks, so no-refill already benefits from early exits and refill provides less improvement.

\section{Experiments}

We evaluate CDB in two settings.
\emph{Offline throughput} measures how fast the engine processes a fixed workload where all requests are known upfront, as in benchmarking or RL.
This sustained load lets us test how much of the estimated maximum speedup CDB can realize and whether the recurrent-step latency model predicts well when refill helps.
\emph{Online serving} replays requests with dynamic arrivals, simulating user queries that arrive at different times, adding queueing effects and varying load.

\subsection{Setup}

We construct three request workloads with different prompt and output lengths: Alpaca (short instructions)~\citep{taori2023alpaca} , ShareGPT (chat requests with similar average prompt and output lengths)~\citep{sharegpt2023}, and ArXiv-Summarization (long, prefill-heavy prompts)~\citep{cohan2018discourse}.
Varying prompt and output lengths tests CDB across different prefill costs, decode durations, and context lengths, all of which affect end-to-end speedup.
We also vary each model's exit threshold to change how often tokens leave the recurrent core and open slots for refill.
For each workload and exit threshold, we set the requests' output length to that of their reference responses and replay exit decisions recorded in advance from the model's exit policy (\cref{app:benchmark-workloads}).

We evaluate CDB on the two main looped LM architecture families, i.e., fully looped models and models with transformer layers in their prelude and coda.
Ouro 1.4B~\citep{zhu2025ouro} is fully looped ($0$-$24$-$0$, $r_{\max}{=}4$): all its transformer layers are looping, with only the token embedding and LM head outside the  recurrent core.
Huginn 3.5B~\citep{geiping2025huginn} has a $2$-$4$-$2$ prelude–core–coda split ($r_{\max}{=}16$), so 50\% of its transformer layers are outside the core.
This contrast tests how boundary stages affect the efficiency of depth-adaptive inference and which architecture design is better suited for it.
Unless stated otherwise, we set the minimum coda batch size to $K{=}1$ for Ouro and $K{=}W/4$ for Huginn, where $W$ is the maximum number of active sequences (e.g., $K{=}16$ at $W{=}64$).

We compare CDB using no-refill and refill against standard token-level CB, which runs $r_{\max}$ loops for every token.
All schedulers use the same codebase, kernels, and admission policy, build on Hugging Face Transformers~\citep{hftransformers}, vLLM~\citep{kwon2023pagedattention}, and SGLang~\citep{zheng2024sglang}.
All experiments run on a single 80\,GB H100 GPU.
See \cref{app:impl} for further details.

\subsection{Offline Throughput}
\label{sec:batch-throughput}

\begin{figure}[t]
  \centering
  \includegraphics[width=1\linewidth]{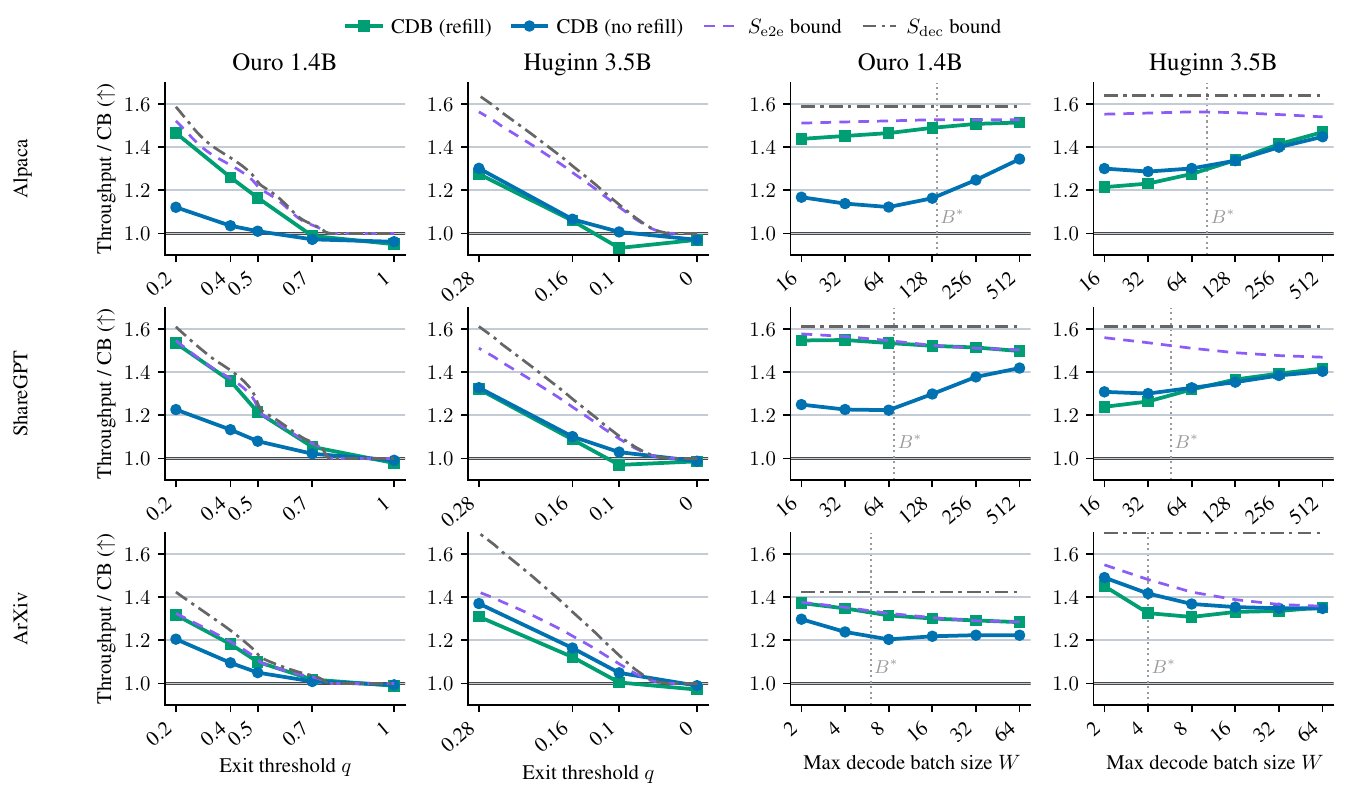}
  \caption{%
    Offline throughput normalized to CB (full depth) for Alpaca (top), ShareGPT (middle), and ArXiv (bottom).
    \emph{Left:} exit threshold sweep for a fixed maximum batch size of 64 (8 for ArXiv).
    \emph{Right:} maximum batch size sweep at $q{=}0.2$ (Ouro) and $q{=}0.28$ (Huginn); dotted lines mark $B^\ast$.
  }
  \label{fig:cb-cdb-throughput}
\end{figure}

We submit all requests at once and report generated tokens per second relative to full-depth CB.
\cref{fig:cb-cdb-throughput} (left) sweeps the exit threshold at a fixed maximum decode batch size of $W{=}64$ ($W{=}8$ for ArXiv), near the saturation batch size $B^\ast$.
The right panels test different batch sizes by sweeping $W$ for a fixed exit threshold.
The dashed curves show the estimated upper bounds on speedup $S_{\mathrm{e2e}}$ and $S_{\mathrm{dec}}$.

\textbf{How close does CDB get to the maximum speedup?}
At the selected thresholds, refill achieves $1.32$ to $1.53{\times}$ the throughput of CB on Ouro and realizes $96$ to $99\%$ of $S_{\mathrm{e2e}}$, nearly matching the estimated maximum for each setting.
On Huginn, no-refill performs best, reaching $1.30$ to $1.37{\times}$ CB and $83$ to $96\%$ of $S_{\mathrm{e2e}}$.
At full depth, CDB has no early exits and therefore no opportunity to outperform CB.
Even with its staged forward pass and additional scheduler interactions, our implementation retains $95$ to $99\%$ of CB throughput, showing that the additional scheduling introduces little overhead.

\textbf{When is refill most effective?}
For Ouro, refill remains within $5\%$ of $S_{\mathrm{e2e}}$ throughout all three batch-size sweeps and consistently outperforms no-refill.
Its advantage is largest around or below $B^\ast$, where decoding is memory bound.
The advantage narrows as larger batches become compute bound, matching the latency model in \cref{sec:cost-model}.
In contrast, for Huginn, no-refill performs better at smaller widths, and the two modes converge as $W$ increases.

\textbf{How much does prefill reduce the available speedup?}
Decode FLOPs alone ($S_{\mathrm{dec}}$) overestimate end-to-end speedup because depth-adaptive decoding does not reduce prefill latency.
Across workloads, prefill accounts for 3.7--36\% of GPU time.
Its share is largest on long-context ArXiv and generally increases with decode batch size as decoding becomes more efficient.
Consequently, $S_{\mathrm{e2e}}$ remains close to $S_{\mathrm{dec}}$ on Alpaca and ShareGPT and falls substantially below it on ArXiv.
Decode still accounts for most latency, supporting our focus on decode-only early exit.

\subsection{Online Serving}
\label{sec:serving}

\begin{figure}[t]
  \centering
  \includegraphics[width=\linewidth]{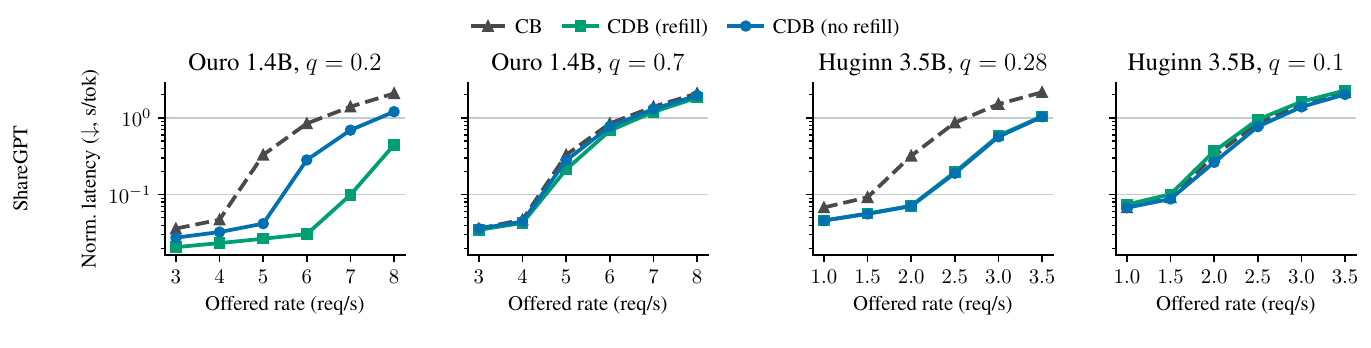}
  \caption{%
    Normalized latency against request rate at two exit thresholds per model on ShareGPT.
    The first and third columns contain thresholds with many exits, and the other columns thresholds with few exits that leave little room for speedup.
    CB is threshold-independent as it always runs full depth.
  }
  \label{fig:serving-rate}
\end{figure}

We replay requests with arrival times sampled from a Poisson process at rate $\lambda$ for a 10-minute window and use a maximum decode batch size of $W{=}64$ ($W{=}8$ for ArXiv).
We report normalized latency (end-to-end time divided by output length), which controls for varying response lengths~\citep{kwon2023pagedattention}.
Once the arrival rate exceeds the serving capacity, requests accumulate in the queue and normalized latency rises sharply.
\cref{fig:serving-rate} shows the ShareGPT results at two exit thresholds per model.
The complete results across workloads and exit thresholds are in \cref{app:serving-sweep}.

\textbf{Do offline throughput gains translate to online serving?}
CDB sustains higher request rates before too many requests overload the system and cause normalized latency to rise, translating the offline throughput gains into greater serving capacity.
For Ouro, refill provides the largest capacity gain, while no-refill adds only about half as much capacity over CB.
For Huginn, refill and no-refill reach the same serving capacity, consistent with the small difference between them in offline throughput.
At thresholds with fewer early exits, the curves approach CB and the capacity advantage disappears.

\subsection{Ablations}
\label{sec:boundary-ablation}

\begin{figure}[htb]
  \centering
  \includegraphics[width=\linewidth]{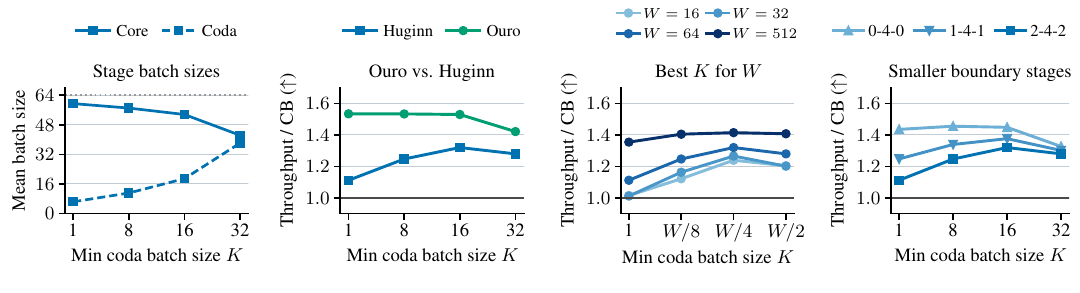}
  \caption{%
    The effects of boundary-stage batching on ShareGPT.
    Blue colors consistently represent Huginn.
    \emph{Left:} effect of $K$ on mean batch size of core and coda on Huginn.
    \emph{Center left:} effect of $K$ on throughput for Ouro and Huginn.
    \emph{Center right:} best $K$ relative to maximum decode batch size $W$ for Huginn.
    \emph{Right:} effect of $K$ on Huginn with different prelude-core-coda layer splits.
  }
  \label{fig:coda-ablation}
\end{figure}

CDB must decide when to interleave coda batches with recurrent-core steps.
Running the coda immediately allows generated tokens to directly refill free core slots.
Waiting for more exits improves coda efficiency at the cost of smaller recurrent batches.
We vary the minimum coda batch size $K$ to measure this trade-off across models, batch sizes, and number of boundary-stage layers (\cref{fig:coda-ablation}).

\textbf{What is the best $K$?}
For Huginn, batching the coda improves throughput until the cost of delayed refill (i.e., recurrent-core batch size shrinks) outweighs the gain in coda efficiency (i.e., coda batch size grows).
Across all decode batch sizes, $K{=}W/4$ best balances these effects.
Ouro has cheap boundary stages, so keeping a large core batch is more important than coda efficiency, so immediate coda execution ($K{=}1$) performs best.
At the largest decode batch sizes, the choice of $K$ matters less because the stages become compute bound, reducing the importance of both coda batching and refill.

\textbf{What if Huginn had smaller boundary stages?}
Removing transformer layers from Huginn's boundary stages makes immediate coda execution more competitive and improves throughput.
For a fully looped $0$-$4$-$0$ split, $K{=}1$ is near optimal, mirroring the behavior of Ouro.
The gap to the maximum speedup also closes, confirming that boundary-stage cost is a central efficiency constraint.

\section{Discussion}

\textbf{Does depth-adaptive decoding result in faster inference?}
CDB realizes up to $99\%$ of the estimated maximum speedup available from the models' early-exit behavior.
However, several conditions must hold to make inference meaningfully faster.
First, more aggressive exit thresholds and missing KV states can reduce accuracy, so the model must be trained to provide enough early exits at the desired accuracy to create compute savings that CDB can use (\cref{app:accuracy}).
Second, recurrent computation must account for a sufficient part of total inference time.
For example, expensive prefill limits the end-to-end speedup, as our prefill-corrected bound shows.
Third, the inference system must turn skipped recurrent computation into lower wall-clock time.
In the memory-bound regime, shrinking the recurrent batch saves FLOPs but little time, making refill important.
Additionally, updating batches between loop steps frequently stalls the GPU, so an asynchronous system is needed.
Overall, our CDB implementation provides the inference method needed, while further inference speedups come from future models that allow more frequent exits at high accuracy.

\textbf{Co-designing looped LMs and inference systems.}
Our results offer concrete guidance for designing looped models that benefit even more from depth-adaptive inference.
Refilling batches between loops can substantially increase inference speed by replacing tokens as they exit.
However, refill needs tokens waiting in the recurrent queue, and tokens only get there after running the boundary stages.
If those stages are costly, they cannot run as often, leaving fewer tokens ready for refill.
This suggests keeping \emph{non-looped stages small} for efficient inference.
However, adding layers outside the recurrent core can improve accuracy~\citep{geiping2025huginn,bae2025mixture}, creating a trade-off between accuracy and inference efficiency.
Furthermore, \emph{more loops} offer finer-grained control over exit depths, but each skipped loop then saves a smaller fraction of total recurrent compute (\cref{sec:cost-model}).
Higher recurrence counts can also reduce accuracy at matched training compute~\citep{schwethelm2026recurrence}.
The KV cache layout creates another co-design requirement.
Early exits leave KV slots for skipped loops unwritten, requiring a policy for missing states. Reusing the last available state avoids recomputation, and looped models can often tolerate this without additional training, though support for KV sharing is not guaranteed.
A \emph{fully shared cache} further reduces the memory footprint and implementation complexity of CDB (\cref{app:kv-cache-sharing}).
Understanding when these layouts work, and designing models that reliably support them, is an important direction for future work.

\textbf{Designing exit gates.}
Existing gates make exit decisions from the latest recurrent state, while asynchronous batch preparation requires knowing exits one loop in advance.
Our lookahead gate provides this early decision without reducing accuracy, but constrains the minimum exit depth to $r_{\min}{=}2$ (\cref{app:early-exit-gating}).
For Ouro and Huginn, excluding exits after only one loop is consistent with their exit behavior, so the constraint does not reduce their realistic speedup.
However, the constraint could reduce the available speedup of models that produce useful outputs after a single loop.
One solution is to predict the complete exit depth before the token enters the recurrent core.
However, making this decision reliable remains difficult~\citep{bae2025mixture}.
A more promising alternative is to predict only the first loop-exit before the recurrent core and then use lookahead decisions for all later loops.
This would allow $r_{\min}{=}1$ without requiring the complete exit depth to be known beforehand.

\textbf{Limitations.}
We evaluate only two looped architectures on a single H100 GPU.
Future looped models may introduce components such as mixture-of-experts that change the costs and scheduling requirements of recurrent and boundary stages.
Our experiments also do not include production serving features such as multi-GPU execution, preemption, prefix caching, or speculative decoding.
These techniques may shift the balance between prefill and decoding or introduce new interactions with depth-adaptive scheduling, so the relative benefit of CDB may differ in a broader serving system.

\section{Conclusion}

We introduced the first efficient inference method for depth-adaptive looped LMs.
Our implementation of continuous depth batching (CDB) decomposes decode steps into separate stage queues, solving the batching problem posed by variable-depth forward passes.
Our method is near-optimal and realizes most of the speedup available from early exits, showing that depth-adaptive decoding can be served efficiently.
We also give new impulses for looped architecture design to improve the compute--accuracy trade-off, e.g., more accurate early exits and less expensive boundary stages.
Overall, co-designing looped architectures and serving systems can lead to substantially faster inference.

\bibliographystyle{iclr2027_conference}
\bibliography{references}

\newpage

\appendix
\crefalias{section}{appendix} 
\crefalias{subsection}{appendix}

\section{Implementation Details}
\label{app:impl}

\subsection{Inference System Overview}
\label{app:system-overview}

\textbf{Engine design.}
Our CB and CDB implementations share the same basic inference design.
Each sequence progresses from \emph{prefill} to autoregressive \emph{decode} and remains active until generation finishes or the scheduler preempts it.
Following SGLang's default setting\footnote{SGLang v0.5.18.}~\citep{zheng2024sglang}, the scheduler forms separate prefill and decode batches based on token, sequence, and KV cache memory limits.
Another common design mixes prefill and decode tokens within the same batch (\enquote{piggybacking}), allowing ongoing generation to advance while new prompts are processed, which reduces long interruptions between decode steps~\citep{agrawal2023sarathi,agrawal2024sarathiserve}. 
We use the phase-separated design to keep decode batches homogeneous, making the throughput effects of early exit easier to isolate.  The CDB method itself does not require phase separation and could also be combined with piggybacking.

\begin{figure}[t]
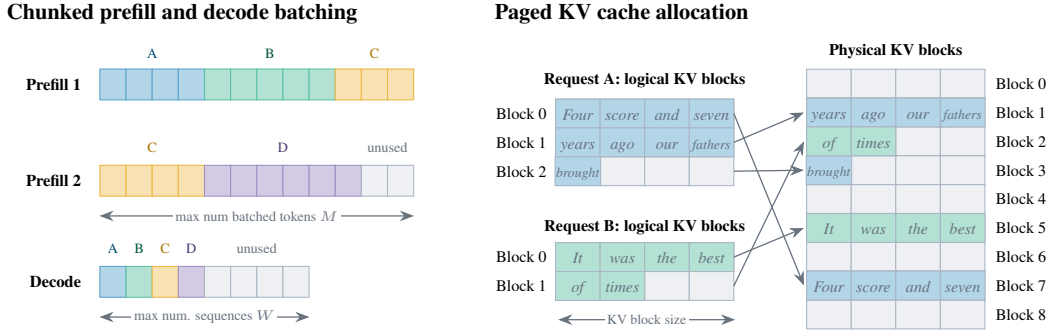

  \centering
  \resizebox{\linewidth}{!}{%
  \begin{tikzpicture}[
    ptitle/.style={font=\fontsize{8.5}{9.2}\selectfont\bfseries, text=black, anchor=west},
    arrow/.style={-{Stealth[length=1.9mm]}, line width=0.5pt, draw=figmuted},
    dimension/.style={{Stealth[length=1.9mm]}-{Stealth[length=1.9mm]}, line width=0.5pt, draw=figmuted}
  ]
    \node[ptitle] at (-1.55,2.3) {Chunked prefill and decode batching};
    \input{figs/implementation/chunked_prefill}
    \node[ptitle] at (5.0,2.3) {Paged KV cache allocation};
    \input{figs/implementation/paged_kv}
  \end{tikzpicture}%
  }
  \caption{%
    \emph{Left:} prompt tokens are packed into prefill batches of at most $M$ tokens, and each active sequence contributes one token to a decode batch of at most $W$ tokens.
    Prompt C is split (chunked) across two prefill batches.
    \emph{Right:} the logical KV blocks of each sequence map to non-contiguous blocks in the shared physical memory pool.
    Blocks are allocated as a sequence grows.
  }
  \label{fig:inference-overview}
\end{figure}

\textbf{Chunked prefill.}
Generation begins with \emph{prefill}, which processes the prompt and writes the resulting attention keys and values to the KV cache.
Since all prompt tokens are known in advance, they can be processed in parallel.
We pack the prompts of multiple requests into a prefill batch containing at most $M$ tokens (\cref{fig:inference-overview}, left).
If a prompt does not fit into the remaining batch budget, we process the part that fits and continue the remaining tokens in the next prefill batch.
This \emph{chunked prefill} bounds the work of each prefill batch and prevents a single long prompt from occupying the GPU in one large forward pass.
Once the final prompt token has been processed, its output produces the first generated token and the sequence enters decode.
Decode is autoregressive: each generated token becomes the input for the next step, so every sequence advances by only one token at a time (\cref{fig:inference-overview}, left).

\textbf{Paged KV cache.}
During generation, each sequence continuously extends its KV cache with the keys and values of newly processed tokens, so its memory footprint grows.
The caches of all active sequences share the limited GPU memory, which bounds the size of the active set and thus the achievable decode batch size.
Output lengths are unknown in advance and memory requirements of sequences may differ substantially.
Reserving space for the maximum possible length of every sequence would admit only a small active set and result in small decode batches.
We use a \emph{paged KV cache}~\citep{kwon2023pagedattention}, which divides the cache into fixed-size blocks and assigns new blocks as a sequence grows (\cref{fig:inference-overview}, right).
This incremental allocation uses the available capacity efficiently, although the cache may still fill before generation finishes, which requires a preemption policy.

\textbf{Preemption.}
The scheduler admits new requests only while sufficient cache capacity is available.
If active sequences nevertheless exhaust the KV-cache capacity because they generate more tokens than expected, the scheduler can remove an active sequence by releasing its KV blocks and returning it to the waiting set.
We implement two preemption policies.
\emph{Offloading} moves the KV state to CPU memory and restores it when the sequence resumes, preserving the values produced during decode.
\emph{Recomputation} discards the KV state and later reconstructs it by processing the prompt and previously generated tokens again through prefill.
Depth-adaptive decoding applies early-exit only during decode, while prefill always runs at full depth.
Tokens that originally exited early are therefore processed again at full depth, so their reconstructed KV values may differ from those produced during decode.
Although this repeats model computation, prefill processes the known tokens in parallel and is much faster per token than autoregressive decode, making cache reconstruction relatively efficient.
Offloading preserves the exact KV state but incurs CPU memory transfers.
We leave a detailed preemption analysis to future work.
We disable preemption in the main experiments so that these costs and state changes do not affect the CB and CDB comparison.

\subsection{Batching and Admission Parameters}
\label{app:batching-admission-parameters}

\cref{tab:inference-config} summarizes the configuration used in our main experiments, and the following paragraphs explain the parameters that control batching and admission.

\begin{table}[t]
\centering
\small
\caption{Inference configuration and parameters used in the main experiments.}
\label{tab:inference-config}
\begin{tabular}{@{}>{\raggedright\arraybackslash}p{0.37\linewidth}p{0.58\linewidth}@{}}
\toprule
Parameter & Setting \\
\midrule
max num sequences $W$ & $64$; $\{2,4,8,16,32,64,128,256,512\}$ in the batch-size sweep \\
max num batched tokens $M$ & $8192$ \\
min free slots $S$ & $\max(1,\lfloor W/8 \rfloor)$ \\
max context length & $16{,}384$ tokens \\
KV-cache capacity & $20{,}480$ blocks ($327{,}680$ token positions) \\
KV block size & $16$ tokens \\
KV-cache headroom $\rho_{\mathrm{KV}}$ & $0\%$ \\
preemption policy & disabled \\
KV-cache layout & shared cache \\
minimum coda batch size $K$ (CDB refill) & $1$ for Ouro; $\lfloor W/4\rfloor$ for Huginn \\
Numerical precision & \texttt{bfloat16} \\
Attention implementation & Paged FlashAttention-3 \\
CUDA graphs & Prefill and decode \\
\bottomrule
\end{tabular}
\end{table}

\textbf{Maximum batch sizes.}
GPU memory limits how many sequences and tokens can be batched together.
Model parameters occupy a fixed part of this memory, while the remaining capacity is primarily shared between the KV cache and intermediate activations.
We therefore use a sequence limit $W$ (\emph{max num sequences}) and a token limit $M$ (\emph{max num batched tokens}).
The KV cache grows with the number and length of active sequences, so $W$ limits how many sequences may retain KV state on the GPU.
Since each active sequence contributes only one token during decode, $W$ also defines the maximum decode batch size (\cref{fig:inference-overview}, left).
In \cref{sec:batch-throughput}, we vary $W$ (and thus the decode batch size) to evaluate CDB in memory- and compute-bound regimes.
During prefill, $M$ limits the total number of prompt tokens batched together.
Since $M>W$ in our experiments, the token limit does not further restrict decode batches.
A larger $M$ allows more prompt tokens to be processed together, which requires more activation memory and may interrupt decode for longer.

\textbf{Prefill admission.}
The active set contains at most $W$ sequences.
When sequences finish, they free slots that can be filled by new sequences.
Because prefill and decode run in separate batches, filling these slots requires a prefill batch, which temporarily pauses decode.
Refilling each slot immediately can produce many prefill batches containing only one or a few prompts.
However, prefill uses the GPU efficiently once a batch contains enough tokens to become compute-bound.
A sufficiently long prompt can reach this regime by itself, so waiting to combine it with other prompts gives little throughput benefit.
Short prompts may not contain enough tokens, making it more efficient to wait for several free slots to combine more prompts.
For example, the short Alpaca prompts generally require more sequences to form a compute-bound prefill batch than the longer ShareGPT prompts.
The \emph{min free slots} parameter $S$ makes the scheduler wait until a chosen number of slots are free before interrupting decode for prefill.
A low value can cause frequent and inefficient prefill interruptions, while a high value makes decode less efficient as the decode batch shrinks with less active sequences.

\textbf{KV-cache headroom.}
The memory footprint of the KV cache depends on the prompt and generated lengths of each active sequence.
The maximum number of sequences $W$ therefore provides only coarse control over memory use.
The \emph{KV-cache headroom} $\rho_{\mathrm{KV}}$ specifies how much of the cache should remain free so that active sequences can add new KV entries as they generate new tokens.
When the free capacity falls below this value, the scheduler stops admitting new sequences and continues decoding the active set.
For example, two sequences with very long prompts may already leave less free cache than the configured headroom, even if $W{=}64$.
The scheduler then keeps the active set at two until one finishes and releases its KV state.
The KV-cache headroom therefore adapts the active-set size to its actual memory use, while $W$ provides a fixed upper bound.

\subsection{Scheduling Algorithms}
\label{app:algorithms}

\cref{alg:no-refill,alg:refill} show pseudocode for CDB's no-refill and refill mode, respectively.
In both algorithms, \textsc{CanPrefill}$(W,S,\rho_{\mathrm{KV}})$ denotes the prefill admission rule described in \cref{app:batching-admission-parameters}.
\textsc{Gate} represents the early-exit gate, which includes checking if a token has reached $r_{\max}$.
The algorithms focus on batch formation and stage ordering and omit asynchronous execution.

\begin{algorithm}[!htb]
  \caption{\emph{No-refill} CDB follows the same phase-separated prefill and decode batching as CB.
  The main difference is that during decode, early exits shrink the recurrent batch without replacement.
  Exited tokens wait until the entire cohort runs the coda together.}
  \label{alg:no-refill}
  \begin{algorithmic}[1]
    \Require prompt queue $Q_{\mathrm{pre}}$, decode queue $Q_{\mathrm{dec}}$, max num batched tokens $M$, max num sequences $W$, min free slots $S$, KV-cache headroom $\rho_{\mathrm{KV}}$, maximum recurrent depth $r_{\max}$
    \Loop
    \While{$\textsc{CanPrefill}(W,S,\rho_{\mathrm{KV}})$}
        \State $B_{\mathrm{pre}} \gets$ pack up to $M$ prompt tokens from $Q_{\mathrm{pre}}$
        \State $x_{\mathrm{new}} \gets$ first generated tokens from $\textsc{FullDepthForward}(B_{\mathrm{pre}})$
        \State push $x_{\mathrm{new}}$ onto $Q_{\mathrm{dec}}$
      \EndWhile
       \State $B_{\mathrm{dec}} \gets Q_{\mathrm{dec}}$
      \State $H \gets \textsc{Prelude}(B_{\mathrm{dec}})$
      \State $E \gets \emptyset$
      \For{$r = 1, \dots, r_{\max}$}
        \State $H \gets \textsc{Core}(H)$
        \State $H_{\mathrm{exit}} \gets \textsc{Gate}(H, r)$
        \State $E \gets E \cup H_{\mathrm{exit}}$;\quad $H \gets H \setminus H_{\mathrm{exit}}$ \Comment{Shrink
  recurrent batch}
        \If{$H = \emptyset$}
          \State \textbf{break} \Comment{All tokens have exited}
        \EndIf
      \EndFor
      \State $x_{\mathrm{next}} \gets \textsc{Coda}(E)$
       \State $Q_{\mathrm{dec}} \gets$ tokens in $x_{\mathrm{next}}$ belonging to unfinished sequences
    \EndLoop
  \end{algorithmic}
\end{algorithm}

\begin{algorithm}[H]
  \caption{\emph{Refill} CDB schedules prefill, recurrent core, and coda independently using the priority rules described in \cref{sec:depth-scheduled-decode}.
  Prelude always runs directly after prefill or coda.
  Tokens that continue looping return to $Q_R$, while exited tokens go to $Q_C$. When $K{=}1$, coda runs immediately after every loop exit, allowing the successor tokens to refill $Q_R$ before the next recurrent core batch.}
  \label{alg:refill}
  \begin{algorithmic}[1]
    \Require prompt queue $Q_{\mathrm{pre}}$, recurrent queue $Q_R$, coda queue $Q_C$, max num batched tokens $M$, max num sequences $W$, min free slots $S$, KV-cache headroom $\rho_{\mathrm{KV}}$, minimum coda batch size $K$, maximum recurrent depth $r_{\max}$
    \Loop
      \If{$|Q_C| \ge K$}
        \State $E \gets Q_C$;\quad $Q_C \gets \emptyset$
        \State $x_{\mathrm{next}} \gets \textsc{Coda}(E)$
        \State $H_{\mathrm{new}} \gets \textsc{Prelude}(x_{\mathrm{next}})$ for unfinished sequences
        \State push $H_{\mathrm{new}}$ onto $Q_R$
      \ElsIf{$\textsc{CanPrefill}(W,S,\rho_{\mathrm{KV}})$}
        \State $B_{\mathrm{pre}} \gets$ pack up to $M$ prompt tokens from $Q_{\mathrm{pre}}$
        \State $x_{\mathrm{new}} \gets$ first generated tokens from $\textsc{FullDepthForward}(B_{\mathrm{pre}})$
        \State push $\textsc{Prelude}(x_{\mathrm{new}})$ onto $Q_R$ for unfinished sequences
      \ElsIf{$Q_R \neq \emptyset$}
        \State $H \gets Q_R$;\quad $Q_R \gets \emptyset$
        \State $H \gets \textsc{Core}(H)$
        \State $H_{\mathrm{exit}} \gets \textsc{Gate}(H, r_{\max})$
        \State push $H_{\mathrm{exit}}$ onto $Q_C$
        \State push $H \setminus H_{\mathrm{exit}}$ onto $Q_R$
      \EndIf
    \EndLoop
  \end{algorithmic}
\end{algorithm}

\subsection{Scheduling Examples}
\label{app:schedule-trace}
\raggedbottom
\newcommand{\tracebatch}[1]{\begingroup\setlength{\fboxsep}{0.5pt}\colorbox{figgreenwash}{\makebox[1.5em][r]{\strut #1}}\endgroup}

\cref{tab:schedule-trace-refill,tab:schedule-trace-ouro-variants} show concrete traces of the scheduling algorithms above.
Each row represents one scheduled stage's batch, where $B$ gives the number of sequences processed by the stage.
For decode stages (prelude, coda, loop) $B$ corresponds to the batch size, while  for prefill the detail column reports the number of batched prompt tokens and thus the prefill batch size.
The queue columns report the queue sizes \emph{after} the stage.
$\mathrm{r}_i{:}n$ in the detail column of loop stages denotes that $n$ tokens execute their $i$-th loop step.
Green boxes highlight the recurrent-core batch sizes.

The traces illustrate how the two modes order the model stages.
With refill, each prefill or coda batch is followed by a prelude, which places the resulting decode tokens directly into $Q_R$.
No-refill instead waits for all prefill steps to finish before starting the decode phase.
With $K{=}1$, coda and prelude run whenever tokens exit, so successor tokens from unfinished sequences can fill the freed recurrent slots before the next loop batch.
Consequently, the recurrent batch remains at its maximum size $W{=}64$ throughout the shown refill traces.
Refill can combine tokens at different loop steps into one forward pass because every step loads the same recurrent-core weights.
For example, stage 14 of the Ouro refill trace combines tokens at their first, second, and fourth loop steps.
Also the shared KV-cache layout makes this easy as every loop step writes and reads to the same KV slot.
No-refill waits until every token has exited, so its recurrent batch only shrinks.
For $K{>}1$, exits accumulate in $Q_C$ and the recurrent batch temporarily shrinks.
A lower $K$ runs coda more frequently and refills the freed slots sooner, trading off a higher recurrent batch size with more coda executions.

\begin{table}[htbp]
  \centering
  \caption{The first 20 stages of refill CDB on ShareGPT with $W{=}64$ and $K{=}1$, using Ouro (left) and Huginn (right).}
  \label{tab:schedule-trace-refill}
  \scriptsize
  \setlength{\tabcolsep}{2.5pt}
  \begin{minipage}[t]{0.49\linewidth}
    \centering
    \begin{tabular}{rlr>{\raggedright\arraybackslash}p{84pt}rrr}
  \toprule
  \multicolumn{7}{c}{\textsc{Ouro: refill} ($K{=}1$)} \\
   & Stage & $B$ & Detail & $Q_R$ & $Q_C$ & $Q_P$ \\
  \midrule
  0 & init &  &  & 0 & 0 & 256 \\
  1 & prefill & 21 & 8192 prompt tokens & 0 & 0 & 235 \\
  2 & prelude & 20 & after prefill & 20 & 0 & 235 \\
  3 & prefill & 33 & 8192 prompt tokens & 20 & 0 & 203 \\
  4 & prelude & 32 & after prefill & 52 & 0 & 203 \\
  5 & prefill & 12 & 1998 prompt tokens & 52 & 0 & 192 \\
  6 & prelude & 12 & after prefill & 64 & 0 & 192 \\
  7 & loop & \tracebatch{64} & $\mathrm{r}_{1}{:}64$ & 64 & 0 & 192 \\
  8 & loop & \tracebatch{64} & $\mathrm{r}_{2}{:}64$; 1 exit & 63 & 1 & 192 \\
  9 & coda & 1 &  & 63 & 0 & 192 \\
  10 & prelude & 1 & after coda & 64 & 0 & 192 \\
  11 & loop & \tracebatch{64} & $\mathrm{r}_{1}{:}1$ $\mathrm{r}_{3}{:}63$; 29 exits & 35 & 29 & 192 \\
  12 & coda & 29 &  & 35 & 0 & 192 \\
  13 & prelude & 29 & after coda & 64 & 0 & 192 \\
  14 & loop & \tracebatch{64} & $\mathrm{r}_{1}{:}29$ $\mathrm{r}_{2}{:}1$ $\mathrm{r}_{4}{:}34$; 35 exits & 29 & 35 & 192 \\
  15 & coda & 35 &  & 29 & 0 & 192 \\
  16 & prelude & 35 & after coda & 64 & 0 & 192 \\
  17 & loop & \tracebatch{64} & $\mathrm{r}_{1}{:}35$ $\mathrm{r}_{2}{:}29$; 9 exits & 55 & 9 & 192 \\
  18 & coda & 9 &  & 55 & 0 & 192 \\
  19 & prelude & 9 & after coda & 64 & 0 & 192 \\
  20 & loop & \tracebatch{64} & $\mathrm{r}_{1}{:}9$ $\mathrm{r}_{2}{:}35$ $\mathrm{r}_{3}{:}20$; 21 exits & 43 & 21 & 192 \\
  \bottomrule
\end{tabular}

  \end{minipage}\hfill
  \begin{minipage}[t]{0.49\linewidth}
    \centering
    \begin{tabular}{rlr>{\raggedright\arraybackslash}p{84pt}rrr}
  \toprule
  \multicolumn{7}{c}{\textsc{Huginn: refill} ($K{=}1$)} \\
   & Stage & $B$ & Detail & $Q_R$ & $Q_C$ & $Q_P$ \\
  \midrule
  0 & init &  &  & 0 & 0 & 256 \\
  1 & prefill & 21 & 8192 prompt tokens & 0 & 0 & 235 \\
  2 & prelude & 20 & after prefill & 20 & 0 & 235 \\
  3 & prefill & 34 & 8192 prompt tokens & 20 & 0 & 202 \\
  4 & prelude & 33 & after prefill & 53 & 0 & 202 \\
  5 & prefill & 11 & 1785 prompt tokens & 53 & 0 & 192 \\
  6 & prelude & 11 & after prefill & 64 & 0 & 192 \\
  7 & loop & \tracebatch{64} & $\mathrm{r}_{1}{:}64$ & 64 & 0 & 192 \\
  8 & loop & \tracebatch{64} & $\mathrm{r}_{2}{:}64$ & 64 & 0 & 192 \\
  9 & loop & \tracebatch{64} & $\mathrm{r}_{3}{:}64$ & 64 & 0 & 192 \\
  10 & loop & \tracebatch{64} & $\mathrm{r}_{4}{:}64$ & 64 & 0 & 192 \\
  11 & loop & \tracebatch{64} & $\mathrm{r}_{5}{:}64$ & 64 & 0 & 192 \\
  12 & loop & \tracebatch{64} & $\mathrm{r}_{6}{:}64$ & 64 & 0 & 192 \\
  13 & loop & \tracebatch{64} & $\mathrm{r}_{7}{:}64$; 7 exits & 57 & 7 & 192 \\
  14 & coda & 7 &  & 57 & 0 & 192 \\
  15 & prelude & 7 & after coda & 64 & 0 & 192 \\
  16 & loop & \tracebatch{64} & $\mathrm{r}_{1}{:}7$ $\mathrm{r}_{8}{:}57$; 22 exits & 42 & 22 & 192 \\
  17 & coda & 22 &  & 42 & 0 & 192 \\
  18 & prelude & 22 & after coda & 64 & 0 & 192 \\
  19 & loop & \tracebatch{64} & $\mathrm{r}_{1}{:}22$ $\mathrm{r}_{2}{:}7$ $\mathrm{r}_{9}{:}35$; 18 exits & 46 & 18 & 192 \\
  20 & coda & 18 &  & 46 & 0 & 192 \\
  \bottomrule
\end{tabular}

  \end{minipage}
\end{table}

\begin{table}[H]
  \centering
  \caption{The first 20 stages for Ouro on ShareGPT with $W{=}64$, using no-refill (left) and refill with $K{=}16$ (right).}
  \label{tab:schedule-trace-ouro-variants}
  \scriptsize
  \setlength{\tabcolsep}{2.5pt}
  \begin{minipage}[t]{0.49\linewidth}
    \centering
    \begin{tabular}{rlr>{\raggedright\arraybackslash}p{84pt}rr}
  \toprule
  \multicolumn{6}{c}{\textsc{Ouro: no-refill}} \\
   & Stage & $B$ & Detail & $Q_D$ & $Q_P$ \\
  \midrule
  0 & init &  &  & 0 & 256 \\
  1 & prefill & 21 & 8192 prompt tokens & 20 & 235 \\
  2 & prefill & 33 & 8192 prompt tokens & 52 & 203 \\
  3 & prefill & 12 & 1998 prompt tokens & 64 & 192 \\
  4 & prelude & 64 &  & 0 & 192 \\
  5 & loop & \tracebatch{64} & $\mathrm{r}_{1}{:}64$ & 0 & 192 \\
  6 & loop & \tracebatch{64} & $\mathrm{r}_{2}{:}64$; 1 exit & 0 & 192 \\
  7 & loop & \tracebatch{63} & $\mathrm{r}_{3}{:}63$; 29 exits & 0 & 192 \\
  8 & loop & \tracebatch{34} & $\mathrm{r}_{4}{:}34$; 34 exits & 0 & 192 \\
  9 & coda & 64 &  & 64 & 192 \\
  10 & prelude & 64 &  & 0 & 192 \\
  11 & loop & \tracebatch{64} & $\mathrm{r}_{1}{:}64$ & 0 & 192 \\
  12 & loop & \tracebatch{64} & $\mathrm{r}_{2}{:}64$; 13 exits & 0 & 192 \\
  13 & loop & \tracebatch{51} & $\mathrm{r}_{3}{:}51$; 39 exits & 0 & 192 \\
  14 & loop & \tracebatch{12} & $\mathrm{r}_{4}{:}12$; 12 exits & 0 & 192 \\
  15 & coda & 64 &  & 64 & 192 \\
  16 & prelude & 64 &  & 0 & 192 \\
  17 & loop & \tracebatch{64} & $\mathrm{r}_{1}{:}64$ & 0 & 192 \\
  18 & loop & \tracebatch{64} & $\mathrm{r}_{2}{:}64$; 17 exits & 0 & 192 \\
  19 & loop & \tracebatch{47} & $\mathrm{r}_{3}{:}47$; 39 exits & 0 & 192 \\
  20 & loop & \tracebatch{8} & $\mathrm{r}_{4}{:}8$; 8 exits & 0 & 192 \\
  \bottomrule
\end{tabular}

  \end{minipage}\hfill
  \begin{minipage}[t]{0.49\linewidth}
    \centering
    \begin{tabular}{rlr>{\raggedright\arraybackslash}p{84pt}rrr}
  \toprule
  \multicolumn{7}{c}{\textsc{Ouro: refill} ($K{=}16$)} \\
   & Stage & $B$ & Detail & $Q_R$ & $Q_C$ & $Q_P$ \\
  \midrule
  0 & init &  &  & 0 & 0 & 256 \\
  1 & prefill & 21 & 8192 prompt tokens & 0 & 0 & 235 \\
  2 & prelude & 20 & after prefill & 20 & 0 & 235 \\
  3 & prefill & 33 & 8192 prompt tokens & 20 & 0 & 203 \\
  4 & prelude & 32 & after prefill & 52 & 0 & 203 \\
  5 & prefill & 12 & 1998 prompt tokens & 52 & 0 & 192 \\
  6 & prelude & 12 & after prefill & 64 & 0 & 192 \\
  7 & loop & \tracebatch{64} & $\mathrm{r}_{1}{:}64$ & 64 & 0 & 192 \\
  8 & loop & \tracebatch{64} & $\mathrm{r}_{2}{:}64$; 1 exit & 63 & 1 & 192 \\
  9 & loop & \tracebatch{63} & $\mathrm{r}_{3}{:}63$; 29 exits & 34 & 30 & 192 \\
  10 & coda & 30 & $\lvert Q_C\rvert\ge 16$ & 34 & 0 & 192 \\
  11 & prelude & 30 & after coda & 64 & 0 & 192 \\
  12 & loop & \tracebatch{64} & $\mathrm{r}_{1}{:}30$ $\mathrm{r}_{4}{:}34$; 34 exits & 30 & 34 & 192 \\
  13 & coda & 34 & $\lvert Q_C\rvert\ge 16$ & 30 & 0 & 192 \\
  14 & prelude & 34 & after coda & 64 & 0 & 192 \\
  15 & loop & \tracebatch{64} & $\mathrm{r}_{1}{:}34$ $\mathrm{r}_{2}{:}30$; 10 exits & 54 & 10 & 192 \\
  16 & loop & \tracebatch{54} & $\mathrm{r}_{2}{:}34$ $\mathrm{r}_{3}{:}20$; 20 exits & 34 & 30 & 192 \\
  17 & coda & 30 & $\lvert Q_C\rvert\ge 16$ & 34 & 0 & 192 \\
  18 & prelude & 30 & after coda & 64 & 0 & 192 \\
  19 & loop & \tracebatch{64} & $\mathrm{r}_{1}{:}30$ $\mathrm{r}_{3}{:}31$ $\mathrm{r}_{4}{:}3$; 25 exits & 39 & 25 & 192 \\
  20 & coda & 25 & $\lvert Q_C\rvert\ge 16$ & 39 & 0 & 192 \\
  \bottomrule
\end{tabular}

  \end{minipage}
\end{table}

\flushbottom
\subsection{Efficient Execution}
\label{app:efficient-execution}

\textbf{CUDA graphs and static shapes.}
A model forward pass consists of many GPU operations, such as matrix multiplications, normalization, and attention.
Many of these operations are executed by separate GPU kernels that the CPU normally launches individually.
The small overhead of each launch accumulates across layers and sequences.
We reduce this launch overhead with \emph{CUDA graphs}, which record a sequence of GPU operations once and replay it with little CPU involvement.
The recorded operations, tensor shapes, and tensor memory addresses must remain fixed between replays, only the values stored in the tensors may change.
We therefore preallocate reusable input and output buffers whose contents are updated for each batch.
To support different batch sizes, we pad batches to a set of supported buckets and capture the corresponding graphs during warm-up.
For CB, each graph captures the complete decode forward pass.
For CDB, we capture separate graphs for the prelude, recurrent core, and coda because the scheduler executes these stages independently.
Both engines capture prefill separately from decode.
In prefill, the attention operation depends on how prompt tokens are divided among sequences, which determines which tokens attend to one another.
Capturing a separate compute graph for every possible combination of prompt lengths would be infeasible.
We therefore break the prefill graph at each attention operation following SGLang.

\textbf{Attention and output projection.}
We use FlashAttention-3 for both prefill and decode~\citep{shah2024flashattention3}.
Prefill uses its variable-length (\emph{varlen}) path, and decode uses its \emph{fast path} for paged KV caches.
The output projection over the vocabulary can also be expensive for large prefill batches.
Only the final token of a completed prompt produces a generated token, so we compute output logits only for these positions instead of for every prompt token.

\section{Recurrent-Step Latency Details}
\label{app:decode-latency}

\begin{figure}[h]
  \centering
  \includegraphics[width=0.32\linewidth]{figs/decode_latency/decode_step_latency_ouro.pdf}\hfill
  \includegraphics[width=0.32\linewidth]{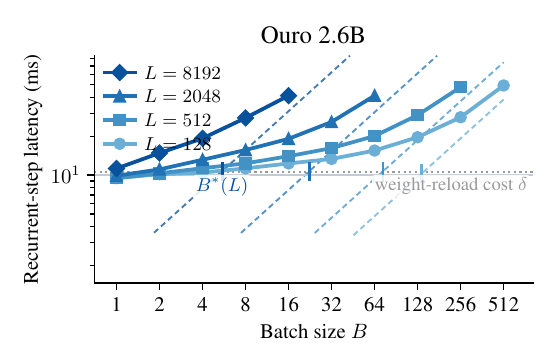}\hfill
  \includegraphics[width=0.32\linewidth]{figs/decode_latency/decode_step_latency_huginn.pdf}
  \caption{%
    Decode latency of a recurrent step versus batch size at context lengths $L$, for Ouro 1.4B (left), Ouro 2.6B (center), and Huginn 3.5B (right).
    The dotted line shows the weight-reload cost $\delta$, with blue ticks marking the saturation batch size $B^*(L)$.
    The blue dashed lines show compute-bound scaling, where latency grows linearly.
  }
  \label{fig:decode-step-latency-app}
\end{figure}

We measure decode latency of a single recurrent step on an 80\,GB H100 GPU through our CB implementation with CUDA graphs and paged FlashAttention-3~\citep{kwon2023pagedattention,shah2024flashattention3}.
For each batch size $B$ and context length $L$, we submit $B$ equal-length requests at full depth (no early exit).
We set up the CB scheduler to complete prefill before any decode begins, so all $B$ requests start decoding together from context length $L$ and the decode batch size is exactly $B$.

For each $(B,L)$, we measure the wall-clock time $\tau$ of a full decode step by averaging over 16 consecutive token generations.
We then separate the fixed boundary cost from the per-step cost by varying $r$.
To isolate the recurrent-step latency, we vary $r \in \{1,2,3,4\}$ for Ouro and $r \in \{2,4,8,12,16\}$ for Huginn and fit
\begin{equation}
    \tau(r) = \tau_0 + \tau_r \cdot r,
\end{equation}
where $\tau_r$ is the recurrent-step latency, and $\tau_0$ is the prelude and coda overhead.
For Ouro, $\tau_0\approx0.4$\,ms across all operating points (embedding and LM head only).
For Huginn, $\tau_0$ ranges from $2.8$\,ms at short context ($L=128$) to $3.3$\,ms at $L=8192$ (due to transformer layers in prelude and coda), about $1.2\tau_r$.
The fit is essentially perfect ($R^2 \ge 0.999$), confirming that decode latency is linear in depth.

We then fit the recurrent-step latency $\tau_r$ with the latency model of \cref{sec:cost-model}, $t(N_c, N_b) = \delta + \gamma\, N_b + \alpha\, N_c$.
At a fixed context length $L$, each of the $N_b = B$ batch tokens reads $L$ KV entries, so $N_c = BL$ and the model reduces to
\begin{equation}
\label{eq:step-cost-fixed-L}
t(B) = \delta + (\gamma + \alpha L)\,B.
\end{equation}
The intercept $\delta$ is the context-independent weight reload cost, $\gamma B$ (mainly) the cost from matrix multiplications, and $\alpha L B$ the KV cache streaming cost.
The saturation batch size is $B^\ast = \delta / (\gamma + \alpha L)$, which falls with $L$ as the KV-cache streaming cost $\alpha L$ grows.

\cref{fig:decode-step-latency-app} shows the fitted curves including the larger Ouro model with 2.6B parameters.
The weight-reload cost $\delta$ depends on layer shape~\citep{anthony2024codesign}.
Huginn's recurrent core has $1.6$B parameters in four wide layers, while Ouro 1.4B has $1.2$B in 24 narrow layers.
Huginn's $\delta \approx 2.3$\,ms is less than half of Ouro 1.4B's $\delta \approx 5.4$\,ms.
Ouro 2.6B doubles the layer count to 48, which approximately doubles both $\delta$ and the slopes, leaving their ratio $B^\ast$ nearly unchanged.
For Ouro 1.4B, $B^\ast$ drops from about 140 at $L=128$ to single digits at $L=8192$.
Huginn follows a similar trajectory from about 80.
Because the measured curves approach the compute-bound asymptote from below without fully reaching it at the largest batch that fits in memory, the fitted slopes mildly underestimate the true per-token cost, so these $B^\ast$ values are mild overestimates.

\section{Benchmark Workloads}
\label{app:benchmark-workloads}

\subsection{Input and Output Tokens}
\label{app:workload-construction}

\begin{figure}[ht]
  \centering
  \includegraphics[width=.9\linewidth]{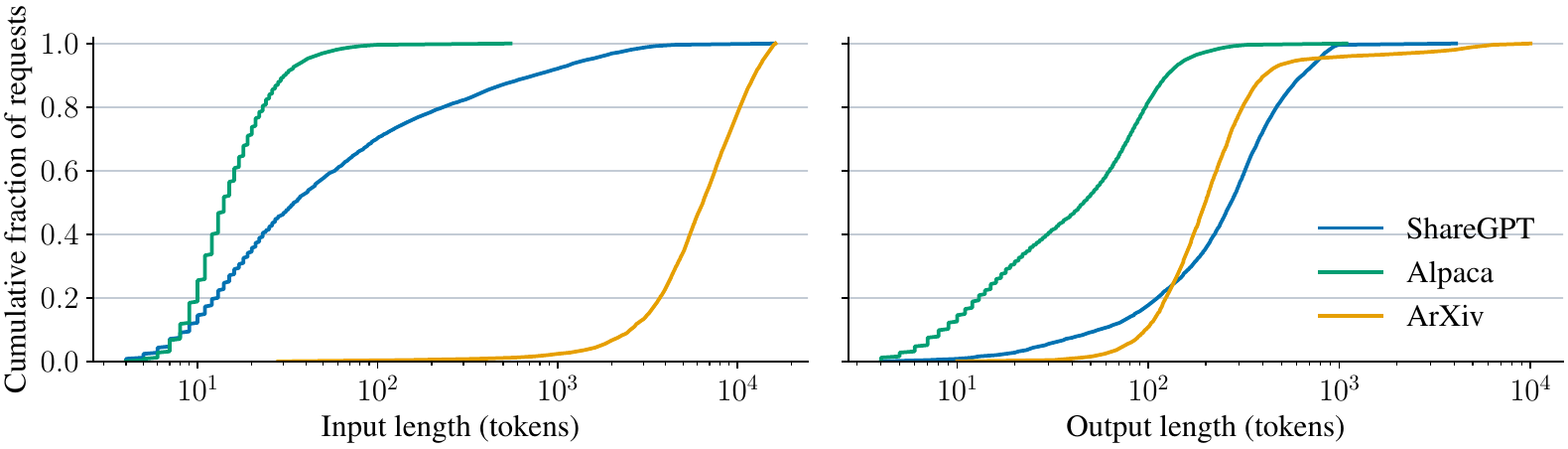}
  \caption{Empirical CDFs of prompt (left) and output (right) token lengths for the three workloads under the Ouro tokenizer.}
  \label{fig:workload-length-distributions}
\end{figure}

In our experiments, we do not let the models generate their own responses, as they may deviate between runs. 
Instead, we take prompt and reference-output lengths from the ShareGPT~\citep{sharegpt2023}, Alpaca~\citep{taori2023alpaca}, and ArXiv-Summarization~\citep{cohan2018discourse} datasets.
From ShareGPT we take one request per conversation by using the first user-assistant pair, so each request is single-turn.
For Alpaca we pair each instruction with its reference output, and for ArXiv we do the same for each article and its corresponding abstract.
We tokenize every prompt and reference output with each model's tokenizer, drop pairs with fewer than four prompt or output tokens, and drop requests whose combined length exceeds the context length of $16{,}384$ tokens.
All tested models replay the same requests in the same order.
Only small differences in token lengths arise from their different tokenizers.

This leaves $56{,}067$ ShareGPT, $49{,}443$ Alpaca, and $11{,}899$ ArXiv requests.
The offline threshold sweeps replay $6{,}000$ ShareGPT, $10{,}000$ Alpaca, and $1{,}000$ ArXiv requests.
The reported throughput is stable at these sample sizes, and running more requests would only increase experiment duration.
For the serving experiments, each point replays a fixed 10-minute Poisson arrival trace, so its request count is determined by the offered request rate.

\cref{fig:workload-length-distributions} shows the resulting input and output token length distributions under the Ouro tokenizer~\citep{zhu2025ouro}.
Alpaca requests are short and tightly concentrated (mean prompt and output lengths of $18$ and $60$ tokens), whereas ShareGPT requests are long and heavy-tailed (mean $267$ and $311$ tokens, with a $99$th prompt-length percentile of $3264$ tokens against Alpaca's $75$).
ArXiv provides a prefill-heavy workload: its mean prompt and output lengths are $6{,}989$ and $370$ tokens, with medians of $6{,}436$ and $199$ tokens.
Its prompts are approximately $26\times$ longer than ShareGPT's on average, while its outputs are only $1.2\times$ longer.
ShareGPT prompts and outputs are in turn on average $14.8\times$ and $5.2\times$ longer than Alpaca's.

\subsection{Early-exit Distribution}
\label{app:exit-recording}

\begin{figure}[h]
  \centering
  \includegraphics[width=.8\linewidth]{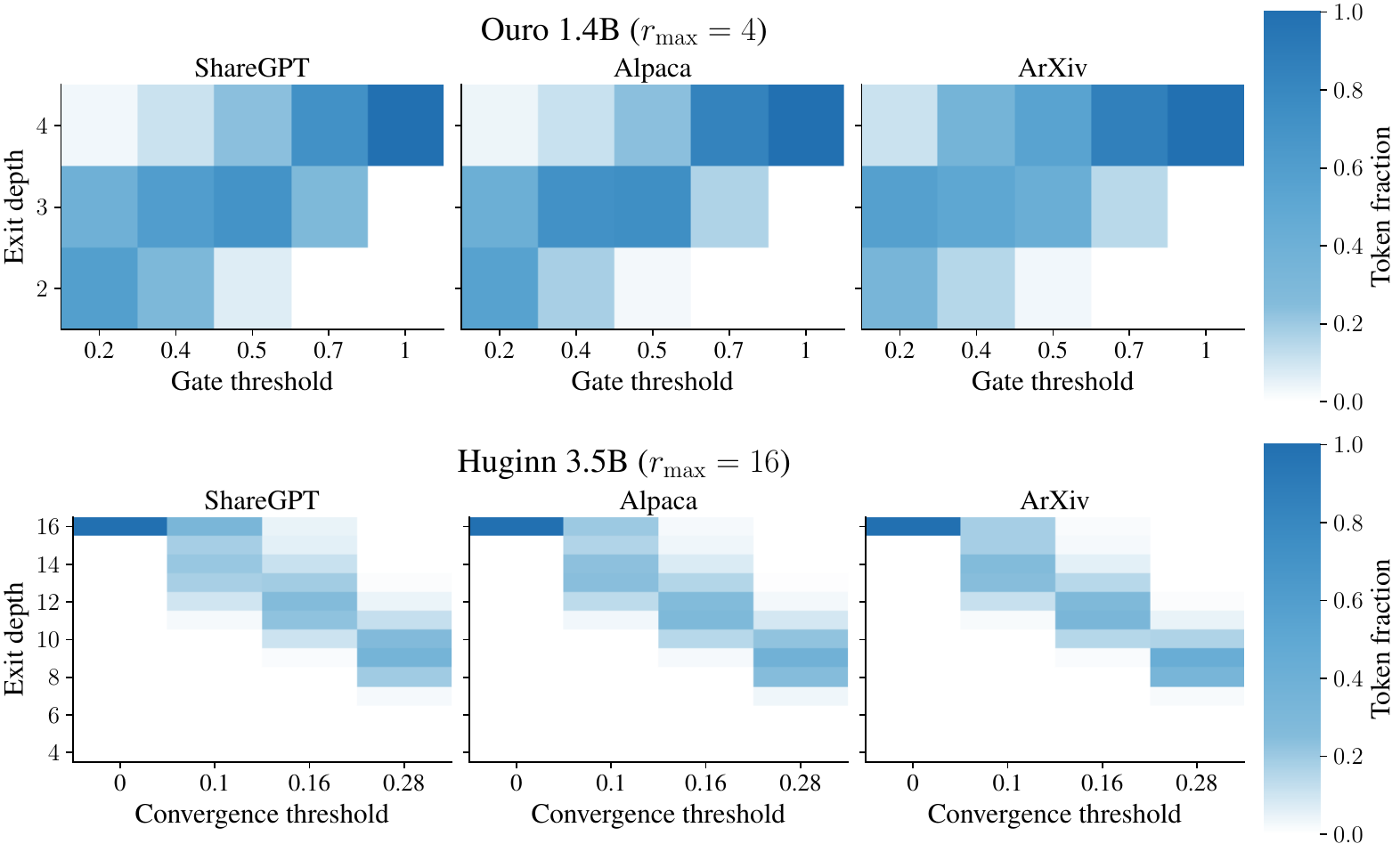}
  \caption{Exit-depth distributions for different gate thresholds. Each cell shows the fraction of tokens exiting at a given depth.}
  \label{fig:workload-exit-distributions}
\end{figure}

Similar to the output tokens, we do not use an active exit gate during our experiments.
To make exit decisions deterministic, we run all requests through the models once in advance and record their per-token exits, which are then replayed during the experiments.
\cref{fig:workload-exit-distributions} shows the resulting exit-depth distributions for different gate thresholds for Ouro and Huginn.
For Ouro, a higher gate threshold makes fewer tokens exit early; for Huginn, a lower convergence threshold has the same effect.
Across all three workloads, the exit-depth distributions remain similar at a given threshold, with only modest workload-dependent shifts.
Huginn's convergence gate spreads exits over a wider range of depths than Ouro's gate.

\section{Complete Online Serving Results}
\label{app:serving-sweep}

\begin{figure}[ht]
  \centering
  \includegraphics[width=\linewidth]{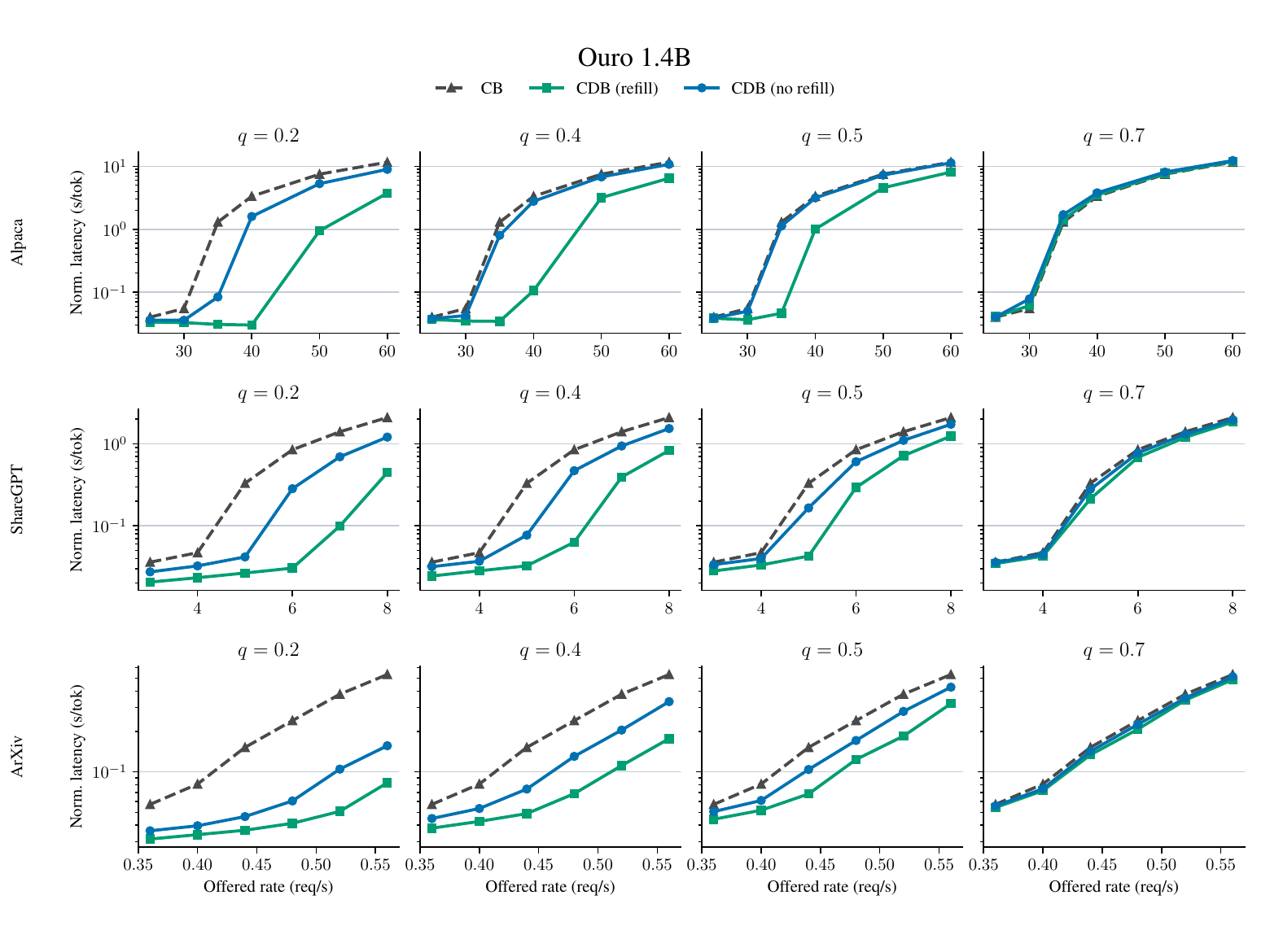}
  \caption{%
    Normalized latency against request rate across all exit thresholds for Ouro on Alpaca, ShareGPT, and ArXiv.
    Columns run from aggressive (left) to conservative (right) thresholds.
    CB is threshold-independent (full depth).%
  }
  \label{fig:serving-sweep-ouro}
\end{figure}

\begin{figure}[ht]
  \centering
  \includegraphics[width=.7\linewidth]{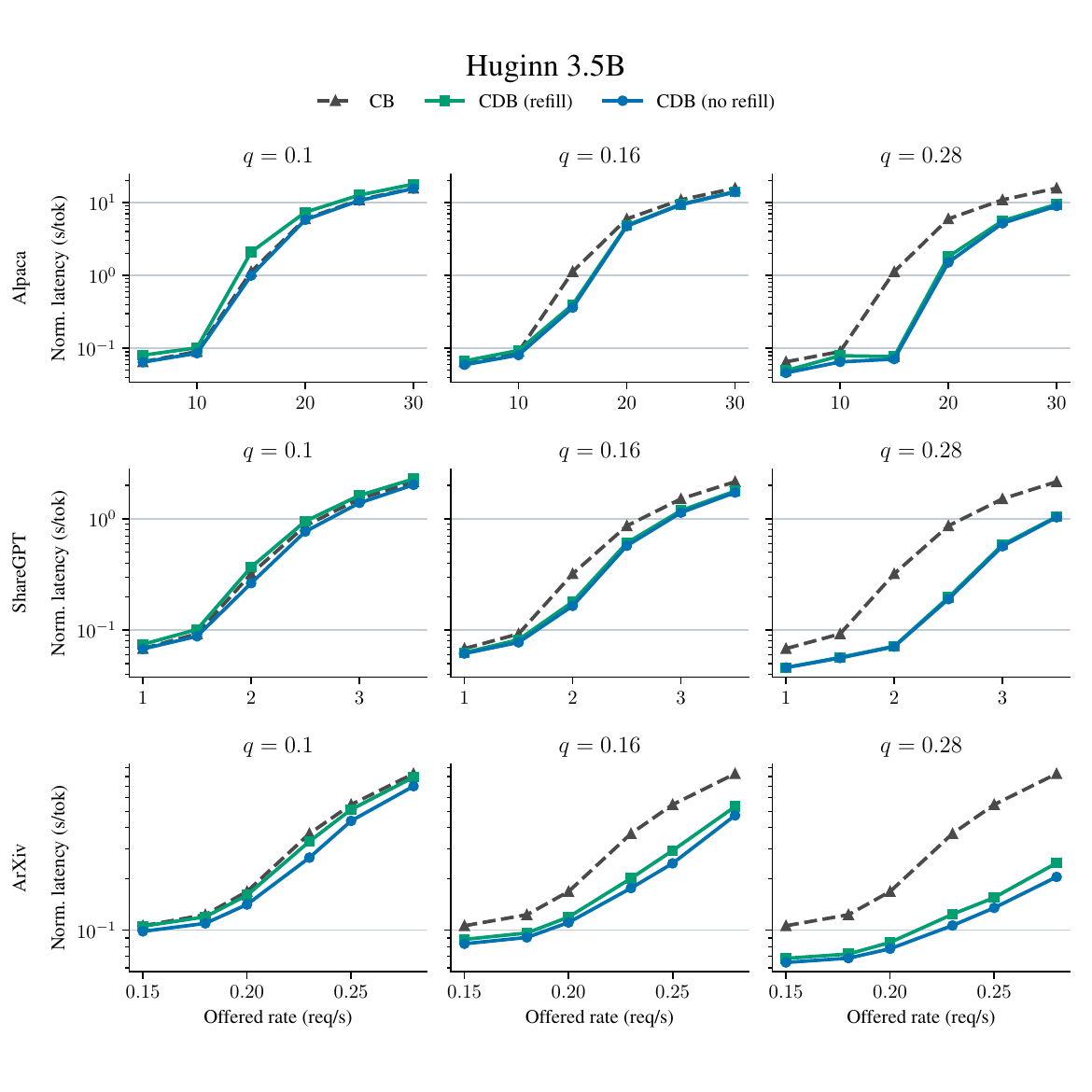}
  \caption{%
    Same as \cref{fig:serving-sweep-ouro} for Huginn, whose exit criterion is inverted: lower thresholds are more conservative, so the columns run from conservative (left) to aggressive (right).%
  }
  \label{fig:serving-sweep-huginn}
\end{figure}

\cref{fig:serving-sweep-ouro,fig:serving-sweep-huginn} extend the ShareGPT comparison of \cref{sec:serving} to Alpaca and ArXiv and more exit thresholds.
The general observations hold for all workloads.

\section{Accuracy-Efficiency Trade-off Ablations}
\label{app:accuracy}

Our work is about maximizing the inference efficiency of looped LMs. This section evaluates how well current models support the techniques we build on. We reimplement the \texttt{gsm8k\_cot} task configuration of the LM Evaluation Harness~\citep{cobbe2021gsm8k,eval-harness} on top of our continuous-batching and continuous-depth-batching implementations, and evaluate Ouro 1.4B~\citep{zhu2025ouro} and Huginn 3.5B~\citep{geiping2025huginn}. Two choices in our setup affect task accuracy: the KV cache policy and the exit gate, both trading off accuracy against throughput and memory.

\subsection{KV Cache Sharing}
\label{app:kv-cache-sharing}

Looped LMs share physical attention layers across iterations, so multiple iterations can also share KV cache slots.
Neither model was trained for this setting, so KV sharing is a zero-shot capability here.
We test how KV cache sharing affects accuracy and memory, evaluating both models at fixed depth.
We compare a \emph{full depth-indexed cache}, where each loop iteration writes to its own KV slot, against two shared layouts: \emph{single-slot}, where every iteration overwrites one shared KV slot per physical layer, and \emph{first-then-shared}, where the first iteration retains a private slot while all subsequent ones share a second.
Writing $s$ for the number of KV slots per recurrent layer, the three layouts use $s = r_{\max}$, $s = 1$, and $s = 2$, respectively.
The total cache thus holds $\ell_0 + \ell_r s$ KV entries per token, where $\ell_0$ denotes the number of attention layers outside the loop (prelude and coda) and $\ell_r$ those inside the recurrent core.
Sharing therefore saves a factor of $\frac{\ell_0 + \ell_r r_{\max}}{\ell_0 + \ell_r s}$ over the depth-indexed cache.

The results in \cref{tab:kv-cache-sharing} show that the Huginn model is insensitive to the KV sharing layout, while Ouro loses accuracy under both layouts, from about 6 points under first-then-shared to a complete collapse under single-slot sharing.
We cannot reproduce the results from the Ouro paper, which reports near-lossless KV sharing~\citep{zhu2025ouro}.
This might be because they share only during decoding and keep a depth-indexed cache for the prompt.
Still, they report a degradation of about 10 accuracy points once the sharing extends to prefill, which is far less than the drop we observe.
Our results are in line with \citet{vendrell2026melt}, who propose integrating KV sharing into Ouro's training as a solution, which then recovers most of the gap.

Surprisingly, Huginn actually performs better under single-slot sharing.
Under a truncated loop (i.e., reducing $r_{\max}$), single-slot sharing retains substantially more accuracy than the full depth-indexed cache: $11.45\%$ against $3.71\%$ at depth $6$ and $25.25\%$ against $12.43\%$ at depth $9$, converging to within about two points at depth $16$ (\cref{tab:kv-depth}).

Overall, sensitivity to KV cache sharing depends on the model, and decreasing it should be a development target, as KV sharing enables not only much simpler CDB (\cref{sec:kv-policy}) but also substantial memory savings: single-slot sharing shrinks the cache $4\times$ for Ouro and $8.5\times$ for Huginn.

\begin{table}[h]
  \centering
  \caption{GSM8K accuracy (flexible match, greedy decoding) and KV cache footprint under different recurrent-depth KV cache layouts. KV slots are counted per physical layer and token. Cache sizes assume \texttt{bfloat16} and are reported per token in KiB and for one full 4096-token sequence in GiB.}
  \label{tab:kv-cache-sharing}
  \small
  \setlength{\tabcolsep}{5pt}
  \begin{threeparttable}
  \begin{tabular}{lcccccccc}
    \toprule
    & \multicolumn{4}{c}{\textsc{Ouro ($r=4$)}} & \multicolumn{4}{c}{\textsc{Huginn ($r=16$)}} \\
    \cmidrule(lr){2-5} \cmidrule(lr){6-9}
    {KV cache layout} & {Slots $s$} & {KiB/tok} & {GiB@4k} & {Accuracy} & {Slots $s$} & {KiB/tok} & {GiB@4k} & {Accuracy} \\
    \midrule
    Full depth-indexed & $4$ & $768$ & $3.00$ & 78.09\% & $16$ & $1402$ & $5.48$ & 30.93\% \\
    Single-slot & $1$ & $192$ & $0.75$ & 0.23\%\tnote{$\dagger$} & $1$ & $165$ & $0.64$ & 32.52\% \\
    First-then-shared & $2$ & $384$ & $1.50$ & 72.40\% & $2$ & $248$ & $0.97$ & 32.83\% \\
    \bottomrule
\end{tabular}

    \begin{tablenotes}[flushleft]
        \footnotesize
        \item[$\dagger$] \citet{zhu2025ouro} report $78.85\%$ for this setting, while \citet{vendrell2026melt} find that Ouro requires specialized fine-tuning to perform well with single slot KV cache sharing.
      \end{tablenotes}
    \end{threeparttable}
\end{table}

\begin{table}[h]
  \centering
  \caption{Huginn 3.5B GSM8K accuracy (flexible match, greedy decoding) under different recurrent-depth KV cache layouts and truncated loops.}
  \label{tab:kv-depth}
  \small
  \setlength{\tabcolsep}{5pt}
  \begin{tabular}{lcccccccc}
    \toprule
    & \multicolumn{8}{c}{Fixed recurrent depth $r$} \\
    \cmidrule(lr){2-9}
    {KV cache layout} & {6} & {7} & {8} & {9} & {10} & {12} & {14} & {16} \\
    \midrule
    Full depth-indexed & 3.71\% & 4.85\% & 8.19\% & 12.43\% & 19.56\% & 27.07\% & 27.37\% & 30.93\% \\
    Single shared & 11.45\% & 15.85\% & 20.39\% & 25.25\% & 28.51\% & 29.80\% & 30.78\% & 32.52\% \\
    First-then-shared & 11.45\% & 14.78\% & 20.02\% & 24.79\% & 27.82\% & 30.25\% & 30.71\% & 32.83\% \\
    \bottomrule
\end{tabular}

\end{table}

\subsection{Early-exit Gating}
\label{app:early-exit-gating}

Looped LMs naturally support test-time compute scaling by varying the loop depth. The simplest strategy is to reduce the fixed depth uniformly for all tokens together. In contrast, depth-adaptive decoding can allocate compute per token, looping less for \enquote{easy} tokens and more for \enquote{hard} ones. We evaluate whether early-exit gating outperforms fixed-depth decoding, and whether the delayed exit decision required by our asynchronous CDB scheduler (\cref{sec:static-async}) degrades accuracy.

\textbf{Setup.}
Ouro includes a trained exit gate, which we use as the baseline. Since our asynchronous CDB scheduler requires the exit decision one step before the actual exit (\cref{sec:static-async}), we train two additional exit gates on the frozen model: a \emph{lookahead gate}, which predicts the exit one step in advance, and a \emph{pre-loop gate}, which predicts the exit before the loop starts. Both are trained via gate distillation on Nemotron-Post-Training-Dataset-v2~\citep{NemotronPostTrainingDatasetV2,nvidia2025nvidianemotronnano2}.
Huginn includes no gate, so we adopt the training-free convergence criterion of \citet{geiping2025huginn}, which exits once the normalized L2 distance between consecutive loop states falls below a threshold.
The threshold direction differs: for Huginn a larger value triggers more exits, for Ouro a smaller one.
For Ouro, we set $r_{\min}{=}2$, as the model is not trained to produce useful outputs after a single iteration~\citep{zhu2025ouro}.
For Huginn, we set $r_{\max}{=}16$ rather than the trained $32$, since accuracy is flat above $16$, and use the shared KV cache layout (single-slot) for better performance (\cref{app:kv-cache-sharing}).

\begin{figure}[ht]
  \centering
  \includegraphics[width=\linewidth]{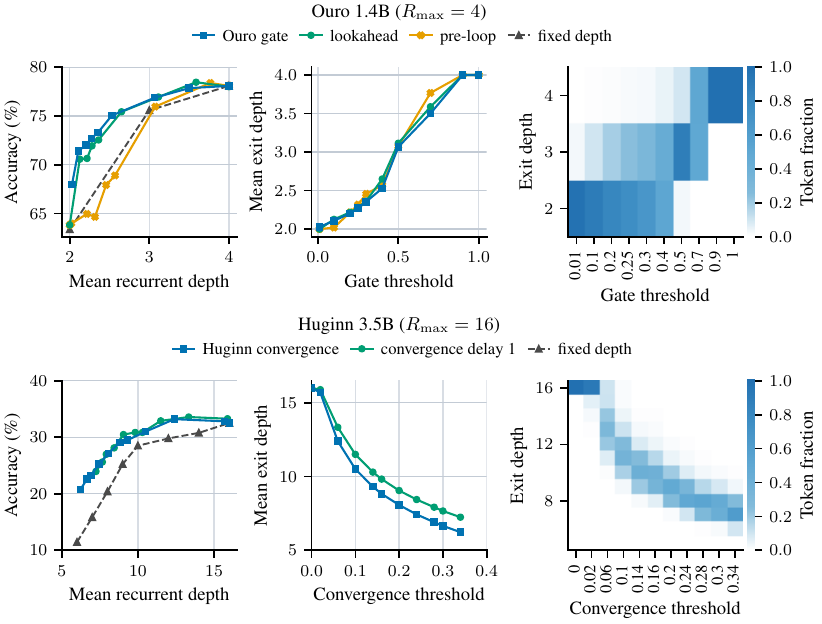}
  \caption{Depth-adaptive decoding against fixed-depth decoding on GSM8K (flexible match, greedy decoding), for Ouro (top) and Huginn (bottom). Each row shows the accuracy--depth frontier (left), the mean exit depth for each threshold (middle), and the exit-depth distribution (right). The distribution panel shows each model's own exit policy (``Ouro gate'' and ``Huginn convergence'').}
  \label{fig:gate-comparison}
\end{figure}

\textbf{Results.}
\cref{fig:gate-comparison} summarizes the results.
Depth-adaptive decoding Pareto-dominates fixed depth, reaching higher accuracy at every mean recurrent depth.
Ouro performs well down to a mean depth of $\approx\!2.5$ and Huginn down to $\approx\!8$.
The exit distributions in the right panel show that Ouro spreads tokens over three of its four depths and Huginn over four to seven of its sixteen.

Regarding delayed exits, the trained lookahead gate closely matches the exit behavior of the teacher gate, achieving approximately the same accuracy and threshold curves.
Thus, the exit can be predicted one step earlier just as well as in the same step, validating our scheduler design.
Predicting the exit before the loop starts would give the scheduler even more flexibility, but the pre-loop gate does not perform well, confirming the observation of \citet{bae2025mixture}.
For Huginn, the delayed exit decision requires only a threshold recalibration and also incurs no accuracy cost.

Overall, gating is effective for both models, and our delayed exits do not degrade accuracy.
However, the minimum loop count remains $2$ for the trained gate and $3$ for the convergence criterion (which compares two consecutive latent states, plus the additional step from the delayed scheduler), limiting the exit distribution and the compute savings.
Ouro and Huginn naturally fit our setup, as they do not exit at those depths, but shallower looped LMs might be more constrained.

\end{document}